\documentclass[10pt,twocolumn,letterpaper]{article}

\usepackage{cvpr}              

\usepackage{amsmath,amsfonts,bm}

\def\eqref#1{equation~\ref{#1}}

\def\1{\bm{1}}

\DeclareMathAlphabet{\mathsfit}{\encodingdefault}{\sfdefault}{m}{sl}
\SetMathAlphabet{\mathsfit}{bold}{\encodingdefault}{\sfdefault}{bx}{n}

\usepackage{amsmath}
\usepackage{amsfonts}
\usepackage{amssymb}
\usepackage{wrapfig}
\usepackage{subcaption}
\usepackage{multirow}
 \usepackage{mathtools} 

\usepackage{verbatim}

\usepackage{anyfontsize}

\usepackage{microtype}
\usepackage{graphicx}
\usepackage{booktabs} 
\usepackage{multirow}
\usepackage{amsmath,amssymb}
\usepackage{booktabs}
\usepackage{caption,subcaption}

\usepackage{xcolor}
\definecolor{mygreen}{HTML}{3cb44b}
\definecolor{skyblue}{HTML}{beffff}
\definecolor{lightgreen}{HTML}{90ee90}

\usepackage{color, colortbl}

\definecolor{emerald}{rgb}{0.31, 0.78, 0.37}

\usepackage{tcolorbox}
\usepackage{enumitem}
\usepackage{colortbl}

\usepackage{xcolor}
\definecolor{mygreen}{HTML}{3cb44b}
\colorlet{myyellow}{green!10!orange!90!}
\makeatletter

\usepackage{tikz}
\usetikzlibrary{arrows,shapes,snakes,automata,backgrounds,fit,petri}
\usepackage{adjustbox}

\newcommand{\RN}[1]{%
	\textup{\lowercase\expandafter{\it \romannumeral#1}}%
}
\usepackage{tabu}

\newcommand{\beq}{\vspace{0mm}\begin{equation}}
\newcommand{\eeq}{\vspace{0mm}\end{equation}}
\newcommand{\beqs}{\vspace{0mm}\begin{eqnarray}}
\newcommand{\eeqs}{\vspace{0mm}\end{eqnarray}}
\newcommand{\barr}{\begin{array}}
\newcommand{\earr}{\end{array}}

\usepackage{color, colortbl}
\definecolor{Gray}{gray}{0.93}

\usepackage{lipsum}

\usepackage{pifont}
\newcommand{\cmark}{\ding{51}}%
\newcommand{\xmark}{\ding{55}}%

\usepackage{makecell}

\usepackage{xcolor,amsmath}
\usepackage[linesnumbered,ruled,vlined]{algorithm2e}
\DontPrintSemicolon

\usepackage{xcolor}
\definecolor{mygreen}{HTML}{3cb44b}

\SetKwComment{Comment}{\color{green!50!black}\# }{}

\SetKwProg{Function}{def}{:}{}

\SetKwProg{For}{for}{:}{}
\SetKwProg{If}{if}{:}{}

\usepackage[dvipsnames]{xcolor}
\usepackage{soul}

\usepackage[numbers, sort&compress]{natbib}

\definecolor{darkred}{RGB}{140, 21, 21}
\definecolor{lightgray}{gray}{0.7}
\definecolor{orange}{HTML}{F58025}

\def\eg{\emph{e.g}\onedot} 
\def\ie{\emph{i.e}\onedot}

\usepackage[T1]{fontenc}

\usepackage{gradient-text}

\newcommand{\tokenvibe}{\textbf{\gradientRGB{VibeToken}{0,76,153}{102,178,255}}}
\newcommand{\tokenvibegen}{\textbf{\gradientRGB{VibeToken-Gen}{0,76,153}{102,178,255}}}

\newcommand{\gradtxtbf}[1]{\textbf{\gradientRGB{#1}{0,76,153}{102,178,255}}}

\usepackage{booktabs, multirow, makecell, colortbl, xcolor, pifont}
\usepackage{array}

\newcommand{\na}{N/A}
\newcommand{\dash}{--}

\newcommand{\yes}{\textcolor{green!60!black}{\cmark}}
\newcommand{\no}{\textcolor{red!65!black}{\xmark}}
\definecolor{headergray}{RGB}{245,245,247}
\definecolor{band}{RGB}{250,248,244}
\definecolor{hilite}{RGB}{255,247,232}  
\usepackage{arydshln}

\definecolor{cvprblue}{rgb}{0.21,0.49,0.74}
\usepackage[pagebackref,breaklinks,colorlinks,allcolors=cvprblue]{hyperref}
\usepackage[accsupp]{axessibility}  

\def\paperID{*****} 
\def\confName{CVPR}
\def\confYear{2026}

\title{\gradtxtbf{VibeToken:} Scaling 1D Image Tokenizers and Autoregressive Models \\for Dynamic Resolution Generations}

\author{Maitreya Patel~$^{1,2,3}$, \quad Jingtao Li~$^{2}$, \quad Weiming Zhuang~$^{2}$, \quad Yezhou Yang~$^{1}$, \quad Lingjuan Lv~$^{2}$\\
\normalfont{$^1$~\textbf{Arizona State University}} \quad
\normalfont{$^2$~\textbf{SonyAI}}\\
}

\begin{document}

\newcommand{\llv}[1]{{\color{orange}[llv: #1]}}
\newcommand{\wm}[1]{{\color{blue}[wm: #1]}}
\twocolumn[{%
\renewcommand\twocolumn[1][]{#1}%
\maketitle

\vspace{-0.5cm}

\centering
\captionsetup{type=figure}
\includegraphics[width=\linewidth]{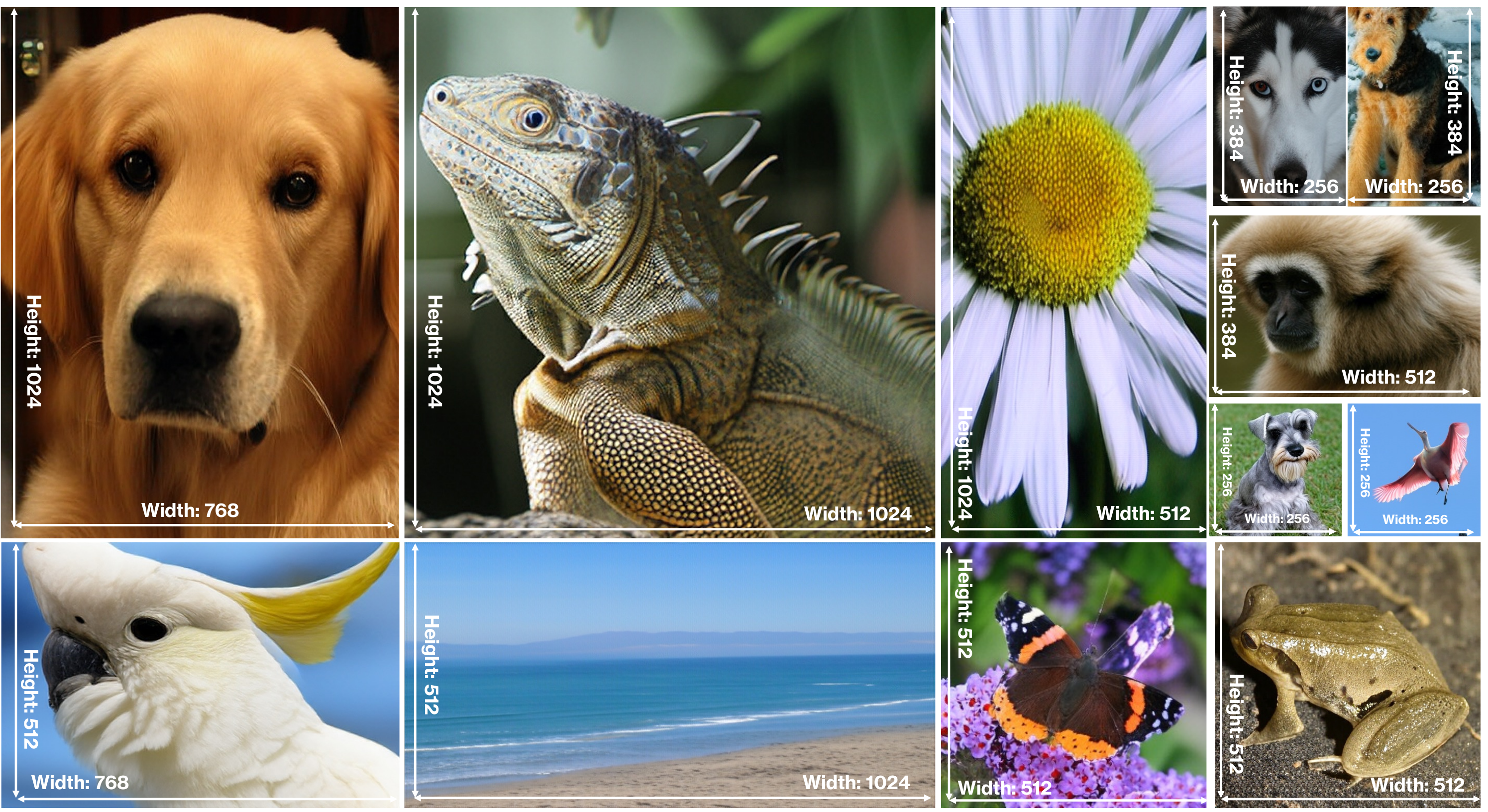}

\captionof{figure}{\textbf{\tokenvibegen\ image generation examples.} A single resolution-generalist visual autoregressive model trained on ImageNet1k that can synthesize arbitrary resolution and aspect ratio images.
The examples shown are from various resolutions between 256$\times$256 and 1024$\times$1024.
\tokenvibegen\ leverages our novel \textit{resolution-agnostic} 1D visual tokenizer, \tokenvibe, that can efficiently encode any resolution into as few as 64 tokens, making our AR model 63$\times$ efficient than the baseline (LlamaGen) on a 1024$\times$1024 resolution.
}
\label{fig:teaser}
\vspace{8pt}
}]

\begin{abstract}

We introduce an efficient, resolution-agnostic autoregressive (AR) image synthesis approach that generalizes to arbitrary resolutions and aspect ratios, narrowing the gap to diffusion models at scale. 
At its core is \tokenvibe, a novel resolution-agnostic 1D Transformer-based image tokenizer that encodes images into a dynamic, user-controllable sequence of 32–256 tokens, achieving a state-of-the-art efficiency and performance trade-off. 
Building on \tokenvibe, we present \tokenvibegen, a class-conditioned AR generator with out-of-the-box support for arbitrary resolutions while requiring significantly fewer compute resources. 
Notably, \tokenvibegen\ synthesizes 1024$\times$1024 images using only 64 tokens and achieves 3.94 gFID; by comparison, a diffusion-based state-of-the-art alternative requires 1,024 tokens and attains 5.87 gFID.
In contrast to fixed-resolution AR models such as LlamaGen—whose inference FLOPs grow quadratically with resolution ($\approx$11T FLOPs at 1024$\times$1024) -- \tokenvibegen\ maintains a constant 179G FLOPs (63.4$\times$ efficient) independent of resolution.
We hope \tokenvibe\ can help unlock the wide adoption of AR visual generative models in production use cases. \textbf{Project page:}
\href{https://github.com/SonyResearch/VibeToken}{https://github.com/SonyResearch/VibeToken}.
\footnotetext[3]{Work partially done while at SonyAI. Maitreya is now at Adobe. Corresponding author:
  \texttt{Lingjuan.Lv@sony.com}.}

\end{abstract}

\section{Introduction}
\label{sec:introduction}

The landscape of visual generative modeling has been transformed by the advent of diffusion and autoregressive (AR) models~\cite{song2021denoising, lipman2023flow, patel2024eclipse, ho2022classifier, yu2022scaling, esser2021taming, sun2024autoregressive, li2024autoregressive, sehwag2025stretching}. 
Diffusion models have emerged as the workhorse for production-scale image generation~\cite{esser2024scaling, saharia2022photorealistic, nichol2021glide, polyak2024movie}. In contrast, AR image synthesis~\cite{sun2024autoregressive, li2024autoregressive}, despite achieving competitive performance, has seen limited deployment in production use cases. 
A key reason is flexibility: diffusion models generalize naturally to arbitrary resolutions and aspect ratios, while AR models struggle to do so. 
As a result, most existing AR literature targets fixed resolutions (e.g., 256$\times$256, 512$\times$512) and fails to adapt to arbitrary resolutions and aspect ratios~\cite{team2025nextstep, tian2024visual, ma2025unitok, wu2024vila, wu2025dc}. 
A common workaround is to append super-resolution modules~\cite{yu2022scaling}, but this introduces additional training complexity and computational cost (e.g., SDXL/Flux-upscaler)~\cite{podell2023sdxl, esser2024scaling}.

\begin{figure*}[t]
  \centering
  \begin{subfigure}[t]{0.32\textwidth}
    \centering
    \includegraphics[width=\linewidth]{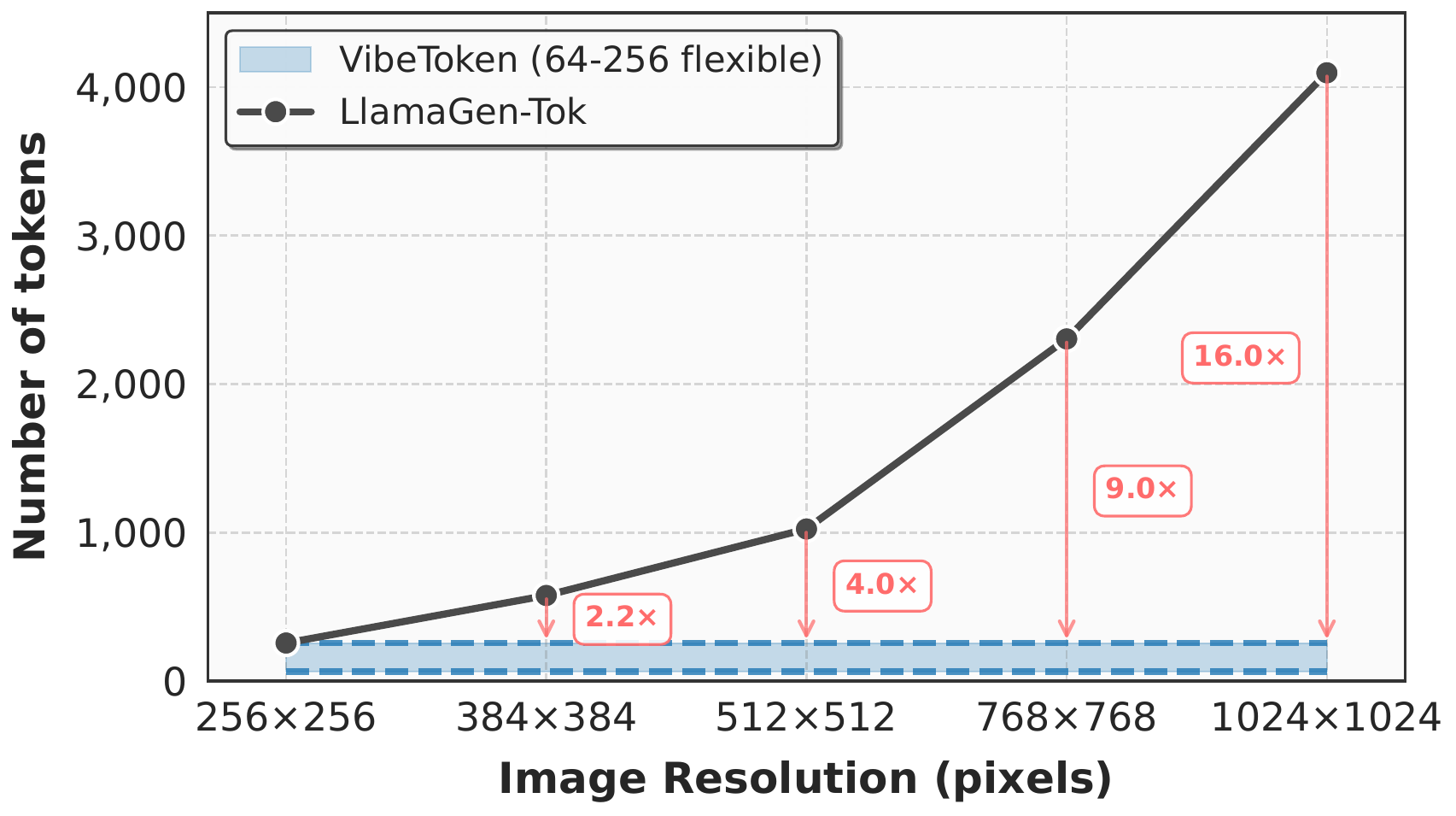}
    \caption{\textbf{VibeToken} keeps token counts to 2-16$\times$ low as resolution increases.}
  \end{subfigure}
  \begin{subfigure}[t]{0.32\textwidth}
    \centering
    \includegraphics[width=\linewidth]{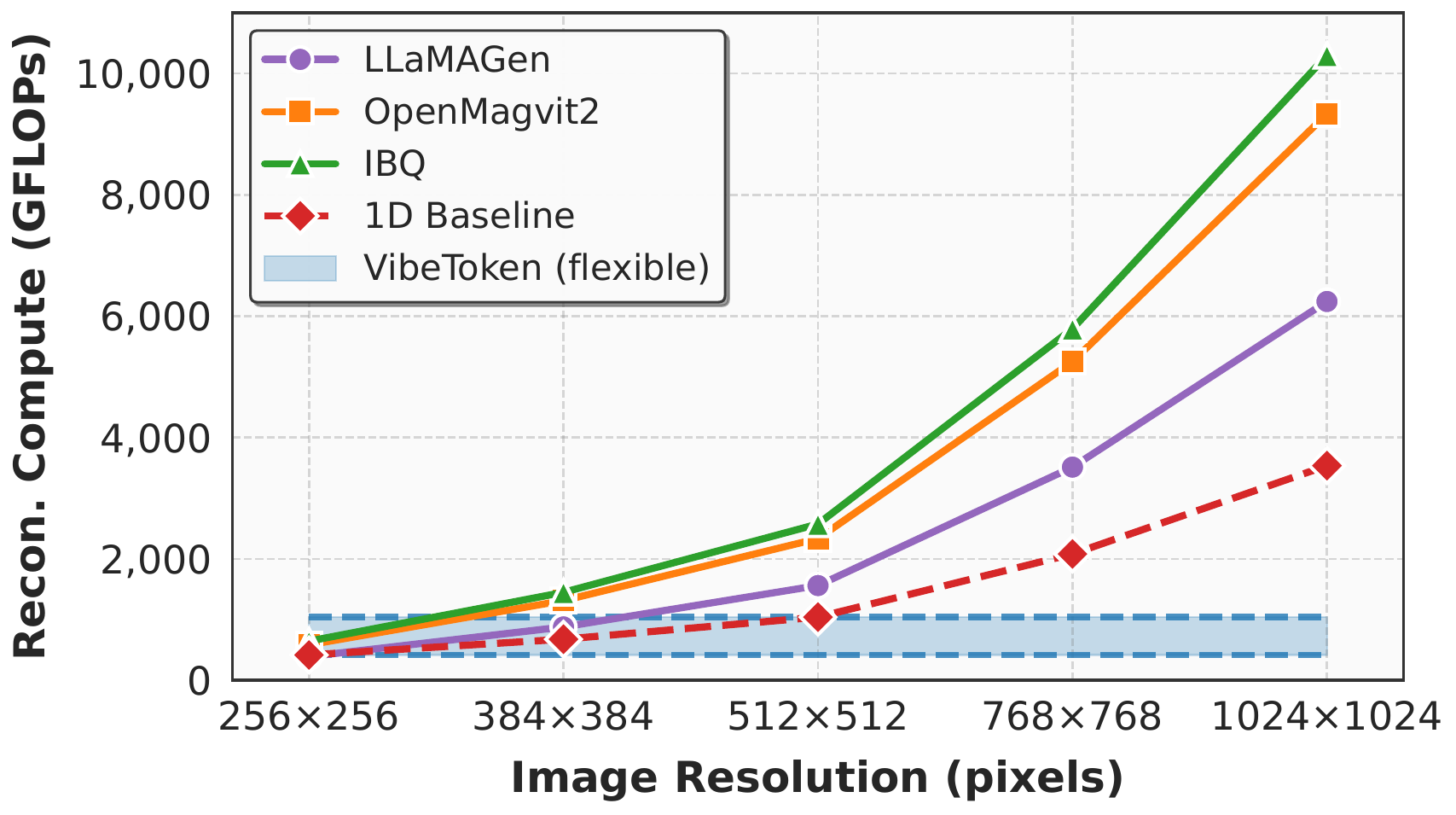}
    \caption{\textbf{VibeToken} keeps FLOPs significantly low and adjustable.}
  \end{subfigure}
  \begin{subfigure}[t]{0.32\textwidth}
    \centering
    \includegraphics[width=\linewidth]{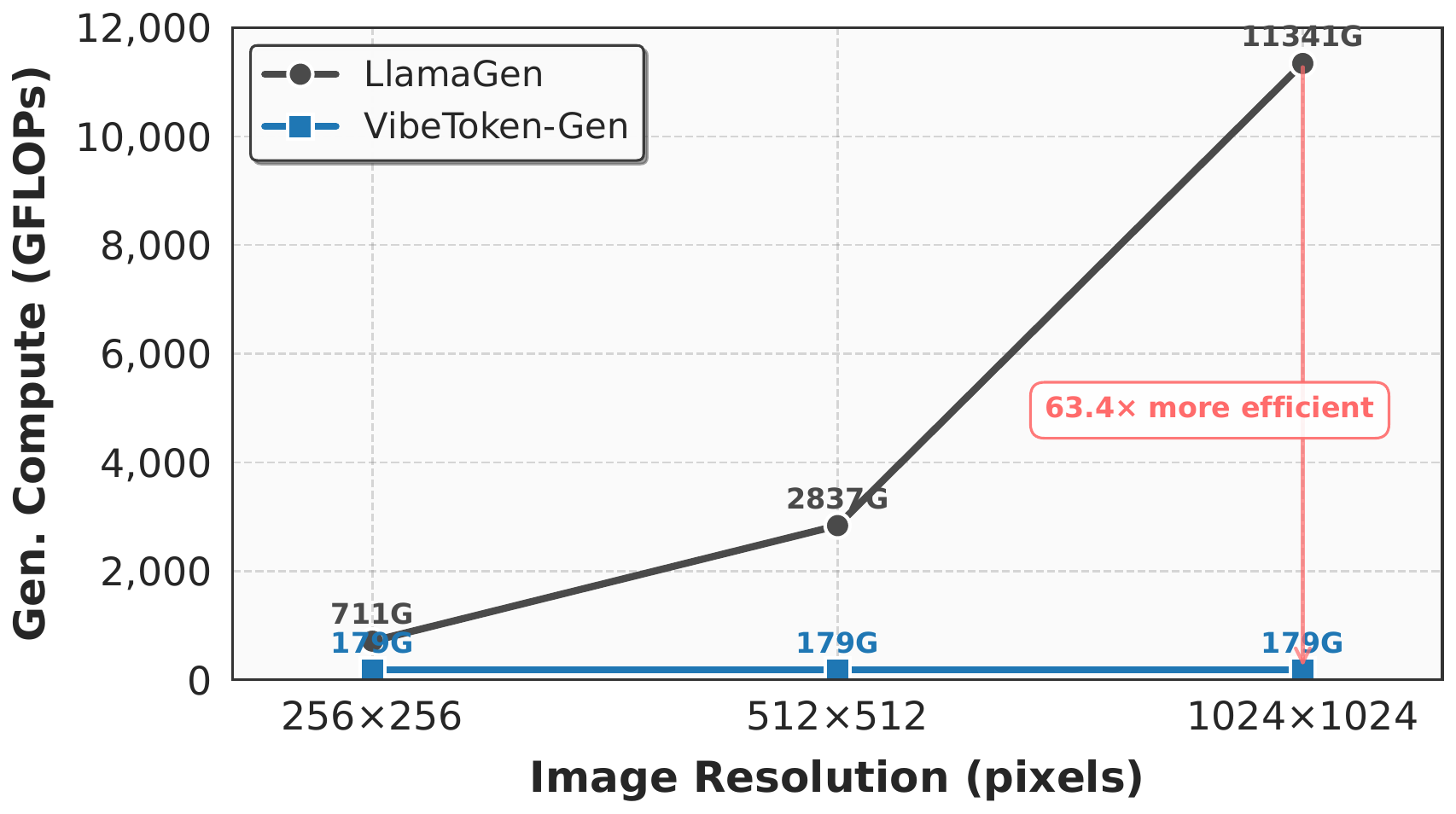}
    \caption{\textbf{VibeToken-Gen} significantly lowers (63$\times$) compute requirements.}
  \end{subfigure}

  \caption{\textbf{Compute comparisons.} Across resolution and metrics, \tokenvibe\ achieves strong efficiency (fewer tokens/FLOPs) compared to 2D baselines.
  This, in turn, leads to \tokenvibegen\ being significantly more efficient with a constant 179 GFLOPs.
  }
  \label{fig:tokenvibe-four-wide}
\end{figure*}

We trace these scalability issues back to the image tokenizer~\cite{yu2022scaling, esser2021taming, sun2024autoregressive, luo2024open, shi2025scalable, zhuang2025argus}. 
As shown in Figure~\ref{fig:tokenvibe-four-wide}, conventional 2D CNN-based tokenizers (e.g., VQGAN~\cite{esser2021taming}) produce a considerable number of tokens that grow linearly with image resolution, which in turn drives AR inference FLOPs to grow nearly quadratically due to self-attention~\cite{esser2021taming, sun2024autoregressive, luo2024open}. 
The need for length generalization in next-token prediction compounds this burden. 
Recent 1D Transformer-based image tokenizers offer stronger compression and partially alleviate scaling challenges, but they still struggle to generalize across resolutions and aspect ratios, unlike their 2D counterparts~\cite{yu2024image, kim2025democratizing, li2024imagefolder, miwa2025one, bachmann2025flextok, chen2025softvq, liu2025detailflow}.

This motivates us to ask the question:
\textit{``Can we encode images of arbitrary resolution into a fixed, small number of tokens and decode them back?''}
If so, an AR generator could be trained at fixed dimensions, while the tokenizer shoulders most of the scaling burden.

In this work, we answer this by proposing novel \emph{resolution-agnostic} image tokenization. 
We scale 1D image tokenizers to support arbitrary resolutions efficiently and introduce \tokenvibe.
It encodes any input resolutions into a dynamic, user-controllable sequence of 32–256 tokens and decodes to any target resolution (input and output resolutions may differ). 
To achieve this, we systematically analyze and scale the architecture, yielding a truly flexible 1D tokenizer via:
(1) dynamic patch embeddings that adapt to input resolutions,
(2) variable image token representations,
(3) efficient positional embeddings robust to resolution and aspect-ratio changes, and
(4) pretraining strategies designed to promote out-of-distribution resolution generalization.
This strategy confers both high compression and strong resolution generalization, and natively supports super-resolution.

Building on \tokenvibe, we present \tokenvibegen, an efficient class-conditioned AR generator and inference pipeline that, for the first time, scales gracefully to arbitrary resolutions and aspect ratios without resolution-specific, compute-intensive training. 
We train \tokenvibegen\ on ImageNet-1k with a training-time token budget comparable to (or smaller than) fixed-resolution AR baselines (e.g., LlamaGen at 256$\times$256). 
We found that 64 tokens are sufficient for \tokenvibegen\ to synthesize arbitrary-resolution images (including 1024$\times$1024).
Crucially, \tokenvibegen\ requires a constant 179~GFLOPs at inference for any resolution, whereas LlamaGen scales to $\approx$11~TFLOPs at 1024$\times$1024.

As a result, \tokenvibegen\ offers substantial efficiency gains over existing AR models and points toward production-grade, resolution-generalist AR modeling.
To summarize, our \textbf{key contributions are following:}
\begin{itemize}[nosep,noitemsep,leftmargin=*]
\item We propose a resolution-agnostic image tokenizer that enables efficient, scalable AR image generation across arbitrary resolutions and aspect ratios.
\item We scale 1D tokenizers to generalize to out-of-distribution resolutions and natively support image super-resolution.
\item \tokenvibe\ achieves state-of-the-art reconstruction among 1D tokenizers while supporting dynamic-length (32–256) encoding.
\item \tokenvibegen\ maintains a constant 179~GFLOPs at inference, whereas prior AR models’ FLOPs grow quadratically with resolution.
\item \tokenvibegen\ is faster (\textbf{2.35$\times$}) and better (\textbf{33\%}) than the diffusion counterpart (NiT) for 1024$\times$1024 synthesis (0.46 s vs 1.08 s), both achieving gFID of 3.94 and 5.87, respectively.
\end{itemize}

\section{Related Works}
\label{sec:related_works}

\paragraph{Autoregressive image generation.}
Visual AR models learn the likelihood of discrete visual tokens, typically in raster-scan order (i.e., row-by-row), through next-token prediction. Early approaches, such as VQGAN+Transformer~\cite{esser2021taming}, ViT-VQGAN~\cite{yu2021vector}, Parti~\cite{yu2022scaling}, RQ-Transformer~\cite{lee2022autoregressive}, and LlamaGen~\cite{li2024autoregressive}, encode images into a discrete latent space and perform AR modeling therein. Despite strong fidelity, scaling is hindered by token growth at higher resolutions and the quadratic attention cost with sequence length. Recent variants improve performance by altering factorization or ordering (e.g., scale-wise AR in VAR~\cite{tian2024visual}; randomized orders in RAR~\cite{yu2025randomized}), achieving speed/quality gains. However, most AR systems are still trained at fixed, relatively low resolutions (256$\,{\times}\,$256 or 512$\,{\times}\,$512), and they struggle to generalize to arbitrary resolutions and aspect ratios due to rapidly increasing token counts and compute.

\paragraph{Image tokenizers.}
Tokenizers determine the sequence length, which in turn affects the feasibility of AR modeling. Classical designs quantize 2D VAE latents with vector codebooks, with many variants exploring quantization strategies such as lookup-free quantization (LFQ)~\cite{sun2024autoregressive} and multi-vector quantization (MVQ)~\cite{ma2025unitok}. 
Recently, DC-AR~\cite{wu2025dc} attempts to improve compression rates up to 32$\times$, but it still requires 1024 tokens for 1024$\times$1024 resolution images.
2D tokenizers often underperform in terms of compression at high resolutions. 
This has motivated Transformer-based 1D tokenizers that learn compact non-spatial latents with stronger compression ratios, including TiTok~\cite{yu2024image}, TA-TiTok~\cite{kim2025democratizing}, One-D-Piece (1DP)~\cite{miwa2025one}, DetailFlow~\cite{liu2025detailflow}, and related methods~\cite{chen2025softvq, bachmann2025flextok, duggal2024adaptive}. 
These methods ease AR training by shortening sequences, but (unlike 2D grid tokenizers) they lack built-in spatial inductive bias and typically assume fixed training resolutions, which limits their generalization across high resolutions and aspect ratios.

\paragraph{Generalization to arbitrary resolutions.}
Resolution scalability is crucial for production use. Diffusion-based models have made notable progress: FiT-v1~\cite{lu2024fit}/FiT-v2~\cite{wang2024fitv2} improve cross-resolution generalization, and NiT~\cite{wang2025native} trains at native resolutions via architectural changes to DiT~\cite{peebles2023scalable}, scaling to arbitrary resolutions within an ImageNet-1k~\cite{deng2009imagenet} compute budget. These advances, however, do not directly transfer to AR pipelines because AR compute scales with token length, which itself grows with image size under conventional tokenizers. Consequently, diffusion remains the default in production. 

Our work aims to bridge this gap for AR by introducing a \emph{resolution-agnostic} 1D tokenizer that maintains a short and controllable token budget across resolutions, allowing an AR generator to operate with a constant inference-time compute budget while supporting arbitrary aspect ratios.

\section{Preliminaries}
\label{sec:prelim}

\paragraph{Autoregressive image generation.}
Let $v\!\in\!\mathbb{R}^{3\times H\times W}$ be an image and let a visual tokenizer $\mathcal{E}$ map it to a sequence of \emph{discrete} tokens $x=\bigl(x_{1},\ldots,x_{T}\bigr)\in[V]^T$, where $V$ is the vocabulary size and $T$ may depend on height $(H)$ and width $(W)$ through $\mathcal{E}$.  
An autoregressive (AR) model factorizes the likelihood via the chain rule:
\begin{equation}
p(x_{1:T}) \;=\; \prod_{t=1}^{T} p_{\theta}\!\left(x_{t}\mid x_{<t}\right).
\end{equation}
In practice, sequence length for arbitrary resolution generalization is handled with an end-of-sequence (EOS) token:
\begin{equation}
p(x_{1:T},\langle\mathrm{eos}\rangle)
\;=\; \prod_{t=1}^{T+1} p_{\theta}\!\left(x_{t}\mid x_{<t}\right),
\quad \text{s.t.}~ x_{T+1}=\langle\mathrm{eos}\rangle.
\end{equation}

The training maximizes the log-likelihood (equivalently, minimizes next-token cross-entropy).
Transformers are the standard choice for $p_{\theta}$. For a sequence of length $T$, self-attention has $\mathcal{O}(T^{2})$ time and memory per layer during training. At inference, AR decoding takes $T$ steps; with KV-caching, the marginal cost of each new token is $\mathcal{O}(T)$, yielding $\mathcal{O}(T^{2})$ total attention cost (without caching it would be $\mathcal{O}(T^{3})$). Since many tokenizers increase $T$ with resolution, the computational burden escalates quickly. For a typical $f$-stride 2D VQ-VAE, $T=(H/f)\cdot (W/f)$; with $f{=}16$, at $256{\times}256$ and $1024{\times}1024$ resolutions, we get  $T{=}256$ and $T{=}4096$, respectively. 
This substantially increases the cost of AR model training and inference, and complicates length generalization.

\paragraph{1D image tokenization.}
Fix patch size $k$. Patchify $v\!\in\!\mathbb{R}^{3\times H\times W}$ into $N=\frac{HW}{k^2}$ patches, project each to $\mathbb{R}^d$, prepend $L$ learned \emph{latent} tokens, and encode:
\[
x^{\mathrm{enc}}=\mathcal{E}_{\theta}(x_0)\in\mathbb{R}^{(N+L)\times d},\qquad
h=x^{\mathrm{enc}}_{N+1:N+L}\in\mathbb{R}^{L\times d}.
\]

Let the codebook be $C=[c_1\cdots c_m]\in\mathbb{R}^{m\times d}$.
The quantizer $Q$ maps each latent to its nearest vector:
\begin{align*}
z = Q(h)\in[m]^{L},
\end{align*}
We apply a reparameterization trick so that the gradients flow to $h$ and $C$.
Then we concatenate quantized latents with $N$ learned masked output tokens $y_{\text{mask}}\!\in\!\mathbb{R}^{N\times d}$:
\[
u=\mathcal{D}_{\phi}\!\bigl([\,C(z)\,\Vert\,y_{\text{mask}}\,]\bigr)\in\mathbb{R}^{(L+N)\times d}.
\]
\emph{Latent positions} $u_{1:L}$ (corresponding to encoded tokens) serve only as conditioning and are \textbf{ignored by the pixel head and loss}.
This yields compact discrete latents for AR modeling, while the decoder predicts pixels exclusively at the masked positions.

\paragraph{Why this formulation struggles to scale?}
Grid-based 2D tokenizers make $T$ grow linearly with increasing resolution. Although 1D Transformer tokenizers often achieve stronger compression, they inherit several bottlenecks akin to AR models:
\begin{itemize}
    \item \textbf{Fixed resolution grid.} The triplet $(k, L, N_{\max})$ is typically fixed at pretraining. Out-of-distribution (OOD) resolutions require (i) ad-hoc positional-embedding interpolation (quality drop), or (ii) separate tokenizers for different resolutions/aspect ratios~\cite{yu2024image}. 
    \item \textbf{Quadratic attention cost;} Each self-attention layer forms a $s\times s$ similarity matrix ($s{=}N{+}L$), incurring $\mathcal{O}(s^{2})$ FLOPs and memory per layer.
    \item \textbf{Rigid downstream generation.} Generators conditioned on discrete tokens inherit fixed or growing token budgets. Increasing resolution increases $T$, making generalization to arbitrary resolutions computationally expensive without explicit retraining.
\end{itemize}

\noindent
These issues motivate a \emph{resolution-agnostic}, compute-controllable tokenizer. In the next section, we introduce \tokenvibe, our first step toward a fully flexible 1D tokenizer that overcomes these obstacles.

\section{Methodology}
\label{sec:method}

\begin{figure}
    \centering
    \includegraphics[width=\linewidth]{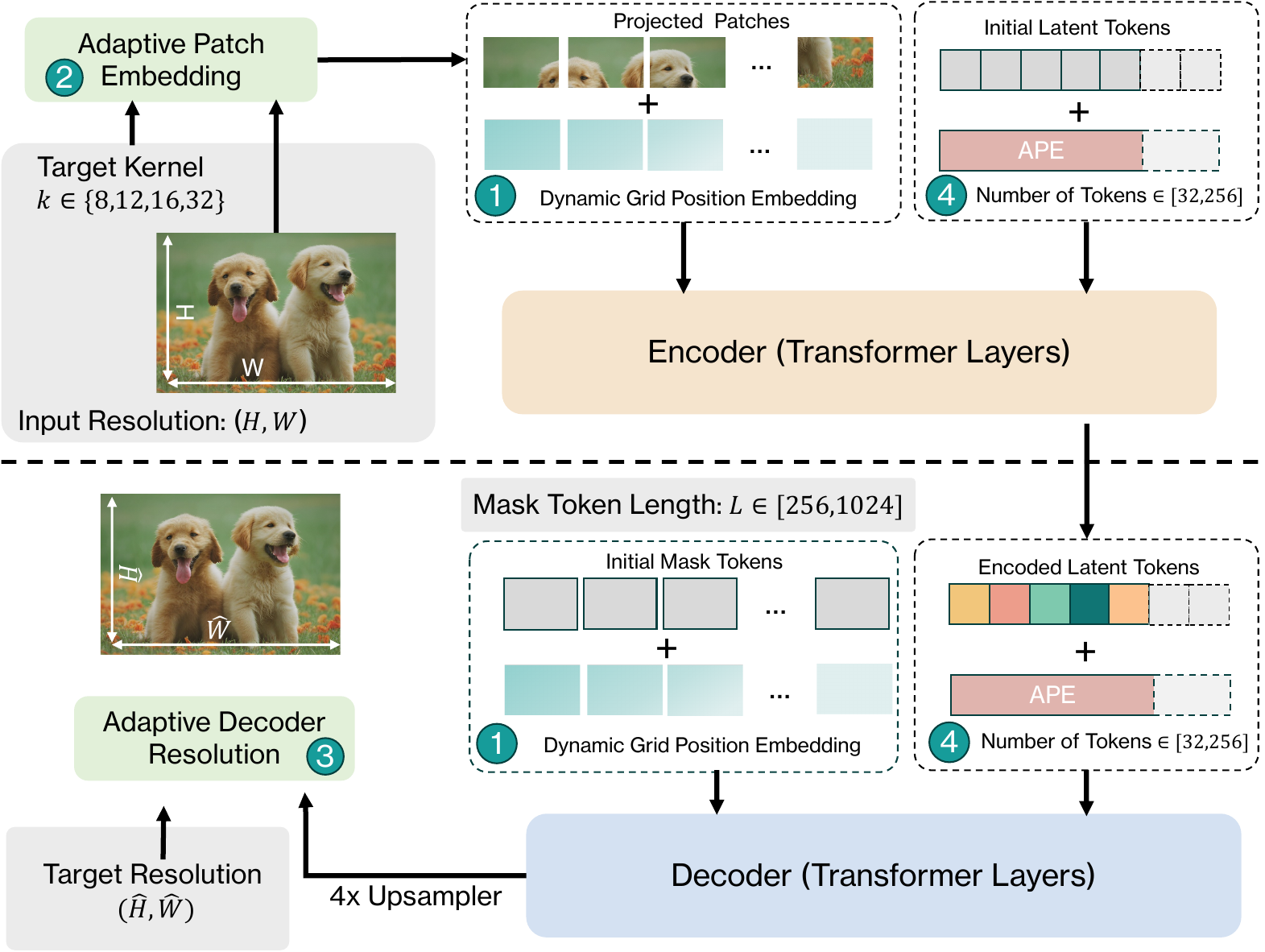}
    \caption{\textbf{Overview of \tokenvibe\ tokenizer.} \tokenvibe\ introduces four key components: 1) Dynamic grid position embedding, 2) Dynamic patch embedding, 3) Adaptive decoder resolution, and 4)  Variable latent token-based resolution-agnostic encoding. These components provide full flexibility to 1D tokenizers, supporting arbitrary resolutions and compute controls.
    }
    \label{fig:architecture}
\end{figure}


We first present \tokenvibe, a 1D tokenizer scaled to learn \emph{resolution-agnostic} visual tokens that generalize to arbitrary resolutions and aspect ratios beyond the training distribution. We then introduce \tokenvibegen, an AR generator (LlamaGen-style) adapted to \tokenvibe\ for efficient high-/arbitrary-resolution synthesis.

\subsection{VibeToken}
\label{sec:tokenvibe}

In Figure~\ref{fig:architecture}, we provide a high-level overview of our tokenizer. 
We focus on positional encoding, patch embedding, decoder adaptations for target resolution control, dynamic-length tokenization, and a training strategy that enforces resolution-agnostic behavior. 

\paragraph{Dynamic grid positional embeddings.}
Varying resolutions imply a varying number of patches; absolute positional embeddings (APE) trained at a single grid often fail to extrapolate. 
Axial RoPE embeddings~\cite{heo2024rotary} help, but can be suboptimal or costly when made layer-wise learnable (see Table~\ref{tab:rope-grid-ablation}). 
We instead adopt a lightweight \emph{dynamic grid position embedding}, inspired by NaViT~\cite{dehghani2023patch} and ViTAR~\cite{fan2024vitar}.
Let $\mathbf{G}\!\in\!\mathbb{R}^{d \times T_H^{\max} \times T_W^{\max}}$ be a learned grid of embeddings (we use $T_H^{\max}{=}T_W^{\max}{=}32$). For an input lattice with
\[
T_H=\Bigl\lceil\tfrac{H}{k_h}\Bigr\rceil,\qquad
T_W=\Bigl\lceil\tfrac{W}{k_w}\Bigr\rceil,
\]
we obtain position codes by differentiable resizing operation (bilinear or bicubic) of $\mathbf{G}$ to $(T_H, T_W)$ as:
\[
\widehat{\mathbf{G}} \;=\; \operatorname{resize}\!\bigl(\mathbf{G};\,T_H,\,T_W\bigr)\;\in\mathbb{R}^{d\times T_H\times T_W},
\]
then flatten to a sequence and add to patch embeddings. Latent (non-spatial) tokens utilize a separate regular APE, as we have a fixed number of latent tokens.
Dynamic grid position embedding preserves inductive bias across resolutions/aspect ratios with negligible overhead and removes the fixed-length constraint of prior 1D tokenizers. 
As noted in Table~\ref{tab:rope-grid-ablation}, it yields a $\sim$33\% FLOPs reduction without loss in quality compared to learnable axial RoPE.

\paragraph{Adaptive patch embedding.}
The length of the sequence $s$ scales with the number of patches. Larger patches reduce computation, but risk detail loss; fixed-$k$ training does not generalize effectively. 
Inspired by FlexiViT~\cite{beyer2023flexivit}, we enable \emph{variable} patch sizes $k\in[k_{\min},k_{\max}]$ at \emph{inference and training}.

Let $p_k\!\in\!\mathbb{R}^{3k^2}$ be a vectorized $k{\times}k$ RGB patch and let $W_{k_{\max}}\!\in\!\mathbb{R}^{d\times 3k_{\max}^2}$ be the \emph{base} projection (\ie kernel of the patch embedding layer). 
We define a \emph{weight-resizing} operator as:
\[
R_{k\leftarrow k_{\max}}\in\mathbb{R}^{3k^2 \times 3k_{\max}^2}
\]
that maps the weights of the $k_{\max}$-grid to the $k$-grid (\eg bilinear interpolation per channel applied $W_{k_{\max}}$). 
Hence, we obtain the differentiable resized kernel of patch embedding as:
\begin{equation*}
W_k \;=\; W_{k_{\max}}\, R_{k\leftarrow k_{\max}},
\qquad
e_{\text{patch}}(p_k)\;=\; W_k\, p_k + b.
\label{eq:adapt-patch}
\end{equation*}
Thus, only $W_{k_{\max}},b$ are learned; all other $W_k$ are derived on-the-fly, avoiding per-$k$ reparameterization and yielding consistent features across $k$. In practice, too large $k$ can affect fine details if $d$ is small; we cap $k_{\max}{=}32$ and \emph{train} over $k\!\in\!\{8,12,16,32\}$ to promote robustness.

\begin{table}[t]
\centering
\scriptsize
\resizebox{\linewidth}{!}{
\begin{tabular}{lccc}
\toprule
\multirow{2}{*}{\textbf{Model}} & \multirow{1}{*}{\textbf{GFLOPs}} &
\multicolumn{2}{c}{\textbf{rFID}$\downarrow$} \\
\cmidrule(lr){3-4}
 &  \textbf{(1024$^2$)} & \textbf{256$^2$} & \textbf{1024$^2$} \\
\midrule
\textsuperscript{{\color{BrickRed}{\textbf{VAE-based small scale study}}}} &  \\[-3pt]
V0 (TiTok-S-128 baseline) & -- & 1.862 & -- \\
 \quad w/ RoPE & 299$^\dagger$ & 2.934 & 280.69 \\
 \quad w/ Learnable RoPE & 445$^\dagger$ & \ul{1.544} & \ul{93.49} \\
\addlinespace[1pt]
 \hdashline
\addlinespace[1pt]
 \quad w/ Dynamic Grid Embedding & 299 & 1.707 & 131.20 \\
 \qquad  + Adaptive Patch Embedding & 90 & \textbf{1.366} & \textbf{5.38} \\
\midrule
\tokenvibe-LL (Learnable RoPE) & 1555$^\dagger$ & 0.46 & 2.57 \\
\tokenvibe-LL (Grid) & 1039 & \textbf{0.40} & \textbf{2.40} \\
\bottomrule
\end{tabular}
}
\caption{\textbf{Ablation of positional embedding and adaptive patch embedding.}
$\dagger$ indicates that FLOPs for RoPE-based models will be slightly higher due to issues in the custom functions calculations.
Ablations are conducted in a VAE setting on a small encoder-decoder model for 200k iterations only.
\tokenvibe-LL represents the final proposed tokenizer, we only perturb position embeddings.
}
\label{tab:rope-grid-ablation}
\end{table}

\paragraph{Adaptive decoder resolution.}
Traditionally, the decoder predicts the fixed values of the $k$ pixels per masked token. Instead, we decode at a \emph{fixed} patch size $k_{\max}=4k$ to an intermediate 4$\times$ high resolution and then introduce an extra downscaling 2D-CNN layer to get the desired target resolution.
Notably, this downsampling layer can be adjusted to get the desired output resolution similar to adaptive patch embedding.
Let the latent length be $L$ and the spatial grid be $(T_H,T_W)$. 
The decoder first predicts patch-wise pixels at resolution:
\[
H' = T_H\,k_{\max},\qquad W' = T_W\,k_{\max},
\]
producing $\tilde v\!\in\!\mathbb{R}^{3\times H' \times W'}$. 
A lightweight learned downscaling 2D-Conv layer $\mathcal{D}_{H,W}$ then maps
\begin{equation}
\hat v \;=\; \mathcal{D}_{H,W}(\tilde v)\;\in\;\mathbb{R}^{3\times H\times W}.
\label{eq:adaptive-dec}
\end{equation}
Similar to adaptive patch embedding, we can control the kernel size of this layer to get the desired target resolution ($H, W$).
This decouples decoding from the input patch size and enables native 4$\times$ super-resolution \emph{without} a separate upscaler.

\paragraph{Dynamic-length tokenization.}
Images vary in complexity; a fixed latent length $L$ is either wasteful or insufficient for ideal compression. 
Tail-token drop (e.g., One-D-Piece~\cite{miwa2025one}) improves this tolerance and introduces dynamic length tokenization; however, it still trains the encoder at a fixed maximum length, potentially creating a quality gap during reconstructions. 
We instead \emph{train} both encoder and decoder on uniform \emph{sampled} lengths $L\!\sim\!\mathcal{P}(L)$ with $L\in[L_{\min},L_{\max}]$ (e.g., 32–256).
Given a target $L$, the encoder produces $L$ latent tokens directly; the decoder consumes exactly $L$, with no padding.
This yields strong quality–compression trade-offs (e.g., 1024$\times$1024 $\rightarrow$ 64 tokens) and seamless compute control at inference, leading to state-of-the-art trade-off between number tokens \textit{vs.} rFID (see Figure~\ref{fig:dynamic_length_recon}).

\begin{table*}
\centering
\resizebox{\linewidth}{!}{
\begin{tabular}{
    >{\raggedright\arraybackslash}p{3.4cm}  
    >{\centering\arraybackslash}p{2.0cm}    
    >{\centering\arraybackslash}p{2.0cm}    
    >{\centering\arraybackslash}p{2.0cm}    
    >{\centering\arraybackslash}p{1.5cm}
    >{\centering\arraybackslash}p{1.5cm}
    >{\centering\arraybackslash}p{1.5cm}
    >{\centering\arraybackslash}p{1.5cm}
}
\toprule
\multirow{2}{*}{\textbf{Models}} &
\multicolumn{2}{c}{\textbf{Dynamic}} &
\multirow{2}{*}{\textbf{Tokens}} &
\multicolumn{4}{c}{\textbf{rFID Performance @ Image Resolutions}} \\
\cmidrule(lr){2-3} \cmidrule(lr){5-8}
 & \textbf{Resolution} & \textbf{Tokens} &  & \textbf{256$^2$} & \textbf{512$^2$} & \textbf{1024$^2$} & \textbf{Arbitrary} \\
\midrule
\textsuperscript{{\color{BrickRed}{\textbf{2D Image Tokenizers}}}} &  \\[-3pt]
LLaMAGen-Tok~\cite{sun2024autoregressive}       & \yes & \no  & 16x$^\ddagger$   & 2.19 & 0.70 & 2.01 & 2.48 \\
Open-MAGVIT-v2~\cite{luo2024open}     & \yes & \no  & 16x$^\ddagger$   & 1.17 & 0.50 & 1.32 & 1.52 \\
IBQ~\cite{shi2025scalable}                & \yes & \no  & 16x$^\ddagger$   & \textbf{0.97} & \textbf{0.40} & \textbf{1.26}  & \textbf{1.42}  \\
DA-HT~\cite{wu2025dc}              & \yes & \no  & 32x$^\ddagger$   & 1.60 & 0.83 & \na  & \na  \\
\midrule
\midrule
\textsuperscript{{\color{BrickRed}{\textbf{1D Image Tokenizers}}}} &  \\[-3pt]
TiTok-B~\cite{yu2024image}            & \no  & \no  & 64/128   & 1.70 & 1.52$^\dagger$ & \dash & \dash \\
VAR~\cite{tian2024visual}                & \no  & \no  & 680      & 0.90 & \dash & \dash & \dash  \\
ImageFolder~\cite{li2024imagefolder}        & \no  & \no  & 286      & 0.80 & \dash & \dash & \dash  \\
DetailFlow~\cite{liu2025detailflow}         & \no  & \no  & 512      & 0.55 & \dash & \dash & \dash  \\
UniTok~\cite{ma2025unitok}             & \no  & \no  & 256      & \textbf{0.33} & \dash & \dash & \dash  \\
Instella~\cite{wang2025instella}           & \yes & \no  & 128      & N/A & 1.32 & N/A & \dash  \\
FlexTok~\cite{bachmann2025flextok}              & \no  & \yes & 1--256   & 1.08 & \dash & \dash & \dash  \\
One-D-Piece-L~\cite{miwa2025one}              & \no  & \yes & 1--256   & 1.08 & \dash & \dash & \dash \\
\hdashline
\rowcolor{hilite}
\tokenvibe--SL      & \yes & \yes & 32--256  & 0.43 & 0.55 & 2.45 & 4.21 \\
\rowcolor{hilite}
\tokenvibe--LL      & \yes & \yes & 32--256  & \ul{0.40} & \textbf{0.51} & \textbf{2.40} & \textbf{3.60} \\
\bottomrule
\end{tabular}
}
\caption{\textbf{Tokenizer comparison across resolutions.}
\yes\ and \no\ indicates whether specific capability is supported or not, \na\ indicates either model not available or failed to reproduce, \dash\ denotes method does not work, and $\dagger$ indicates resolution-specific (independently) trained model. Importantly, $\ddagger$ represents the compression ratios ($f$). Hence, token count ($(H\cdot W)/f^2$) varies \textit{w.r.t.} target resolutions.}

\label{tab:tokenizer-comparison}
\end{table*}


\begin{table*}[t]
\centering
\scriptsize

\resizebox{\textwidth}{!}{
\begin{tabular}{lcccccccccc}
\toprule
\multirow{3}{*}{\textbf{Model}} & \multirow{3}{*}{\textbf{Type}} &
\multicolumn{9}{c}{\textbf{gFID at arbitrary low resolutions}} \\
\cmidrule(lr){3-11}
& & \textbf{1:1} & \textbf{2:3} & \textbf{3:2} & \textbf{3:4} & \textbf{4:3} & \textbf{1:2} & \textbf{2:1} & \textbf{1:1} & \multirow{2}{*}{\textbf{Average}} \\
&  & 256$\times$256 & 256$\times$368 & 368$\times$256 & 368$\times$512 & 512$\times$368 & 256$\times$512 & 512$\times$256 & 512$\times$512 & \\
\midrule
EDM2\text{-}L~\cite{karras2024analyzing}  &     \multirow{3}{*}{Diff.}         & 70.19 &36.29 &32.70 &\ul{3.29} &\ul{3.19} &18.11 &14.30 &\ul{1.92} &22.50 \\
FiTv2\text{-}XL~\cite{wang2024fitv2} &            & \textbf{2.26} &6.28 &6.30 &155.97 &176.66 &27.31 &35.36 &259.11 &83.66 \\
NiT\text{-}XL~\cite{wang2025native}     &           & \ul{2.27} &\textbf{4.28} &\textbf{4.00} &\textbf{2.81} &\textbf{3.00} &\textbf{9.60} &\textbf{6.04} &\textbf{1.80} &\textbf{4.22} \\
\hdashline
\rowcolor{hilite}
\tokenvibegen\ (B) &   & 7.62 &9.19 &7.87 &7.63 &7.46 &15.83 &10.08 &7.45 &9.14 \\
\rowcolor{hilite}
\tokenvibegen\ (XXL) & \multirow{-2}{*}{AR}  & 3.62 &\ul{5.60} &\ul{4.22} &4.00 &3.78 &\ul{12.47} &\ul{7.76} &3.60 &\ul{5.63} \\
\bottomrule
\end{tabular}
}

\vspace{4pt} 

\resizebox{\textwidth}{!}{ 
\begin{tabular}{lcccccccccc}
\toprule
\multirow{3}{*}{\textbf{Model}} & \multirow{3}{*}{\textbf{Type}} &
\multicolumn{9}{c}{\textbf{gFID at higher resolutions}} \\
\cmidrule(lr){3-11}
& & \textbf{1:1} & \textbf{2:3} & \textbf{3:2} & \textbf{3:4} & \textbf{4:3} & \textbf{1:2} & \textbf{2:1} & \textbf{1:1} & \multirow{2}{*}{\textbf{Average}} \\
& & 512$\times$512 & 512$\times$768 & 768$\times$512 & 768$\times$1024 & 1024$\times$768 & 512$\times$1024 & 1024$\times$512 & 1024$\times$1024 & \\
\midrule
EDM2\text{-}L~\cite{karras2024analyzing}   &     \multirow{3}{*}{Diff.}                           & \ul{1.92} &5.62 &6.04 &31.03 &39.54 &17.54 &19.87 &64.32 &23.23 \\
FiTv2\text{-}XL (highres)$^\dagger$~\cite{wang2024fitv2}    &                         & 2.93 &20.61 &29.49 &155.97$^*$ &176.66$^*$ &163.91 &185.89 &259.11$^*$ &124.32 \\
NiT\text{-}XL~\cite{wang2025native}   &                            & \textbf{1.80} &\textbf{4.45} &\ul{4.47} &\ul{4.89} &\ul{5.14} &\ul{12.23} &\ul{9.57} &\ul{5.87} &\ul{6.05} \\
\hdashline
\rowcolor{hilite}
\tokenvibegen\ (B)\footnotemark[3]  &         & 7.62 &8.97 &7.64 &7.57 &7.46 &15.23 &10.07 &7.36 &8.99 \\
\rowcolor{hilite}
\tokenvibegen\ (XXL)\footnotemark[3]  &  \multirow{-2}{*}{AR}  & 3.69 &\ul{5.32} &\textbf{4.01} &\textbf{4.05} &\textbf{3.84} &\textbf{11.91} &\textbf{7.88} &\textbf{3.54} &\textbf{5.53} \\
\bottomrule
\end{tabular}
}
\caption{\textbf{gFID across aspect ratios and resolutions (single view).}
Top: low resolutions (256–512); bottom: high (512–1024). The first header line shows the aspect ratio, and the second line lists the corresponding pixel resolution.
$*$ results are borrowed from low-resolution experiments due to the days of inference computing requirements.
$\dagger$ FiT-v2 has high resolution trained separately for 512$\times$512 and plus.
$\ddagger$ \tokenvibegen\ utilizes the native super resolution of our tokenizer.
}
\label{tab:gfid-resolutions-combined}
\end{table*}

\paragraph{Training strategy.}
We train on images between 256$\times$256 and 512$\times$512 across aspect ratios $\{1\!:\!1,\,1\!:\!2,\,2\!:\!1,\,2\!:\!3,\,3\!:\!2\}$, sampling patch sizes $k\in\{8,12,16,32\}$ such that the number of spatial tokens $N\!\leq\!1024$. 
For each sample, we draw an \emph{input} resolution $(H_{\text{in}},W_{\text{in}})$ and an independent \emph{target} resolution $(H_{\text{out}},W_{\text{out}})$, training the model to reconstruct at $(H_{\text{out}},W_{\text{out}})$ from $(H_{\text{in}},W_{\text{in}})$. 
We also sample the latent length uniformly $L\in[32,256]$. 
For quantization, we adopt a multi-codebook VQ (MVQ) variant compatible with our variable-length setting; implementation details and codebook updates are provided in the Appendix.%
\footnote{Briefly, MVQ improves high-compression performance by distributing capacity across multiple codebooks.}

\paragraph{Summary.}
The combination of Dynamic-APE, adaptive patch embedding, adaptive decoder resolution, and dynamic-length training yields a \emph{resolution-agnostic} tokenizer with strong compression and low overhead.

\subsection{VibeToken-Gen}
\label{sec:tokenvibe-gen}
\tokenvibe\ provides small \textit{resolution-agnostic} discrete token sequences with controllable length $L$. To test its downstream image generation using \tokenvibe\, a natural choice would be to adapt a LlamaGen-style AR model (UniTok head with MVQ compatibility) to operate on these discrete tokens while keeping training compute comparable to a fixed 256$\times$256 resolution baseline.

\paragraph{Conditioning on target resolution/aspect ratio.}
Although \tokenvibe\ can decode to any $(H,W)$, it can exhibit stretching artifacts, such as resizing a square image into a vertical/horizontal shape. 
We therefore condition the AR model on the desired $(H,W)$ alongside the class label $y$, and the rest of the AR stack is unchanged:
\begin{align*}
& c = \bigl[\,\operatorname{emb}(y) + \operatorname{MLP}((H,W)/\beta)\,\bigr],
\end{align*}
where $emb: \mathcal{R}^1 \rightarrow \mathcal{R}^d$ and $MLP: \mathcal{R}^2 \rightarrow \mathcal{R}^d$ are projection layers, while $\beta=1536$ is normalization constant.

\paragraph{Training and inference.}
We train on the same resolution/aspect ratio mix as \tokenvibe, sampling latent lengths $L\in\{64,128,256\}$ with class conditioning. Because $L$ is small (e.g., $\leq\!256$ for up to 1024$^2$) and independent of $(H,W)$, \tokenvibegen\ maintains a \emph{constant} inference-time FLOPs budget across resolutions; compute is governed primarily by $L$ and the AR depth, \textbf{not by target resolution}.

\section{Experimental Results}
\label{sec:exp}

We detail implementation and evaluation results for our tokenizer and the class-conditioned autoregressive generator. 
We report reconstruction and generation performance across resolutions and analyze efficiency.

\captionsetup[subtable]{position=b,font=footnotesize,labelfont=bf,justification=raggedright,singlelinecheck=false,skip=2pt}

\begin{table*}[t]
\centering
\setlength{\tabcolsep}{3pt}

\subcaptionbox{Impact of token length.\label{tab:a}}[0.22\linewidth]{%
\centering\scriptsize
\begin{tabular}{lcc}
\toprule
\multirow{2}{*}{\textbf{Tokens}} & \multicolumn{2}{c}{\textbf{gFID}} \\
\cmidrule(lr){2-3}
 & \textbf{w/o cfg} & \textbf{w/ cfg} \\
\midrule
256 & 39.32 &  9.81 \\
128 & 37.68 &  9.02 \\
 64 & \textbf{33.15} & \textbf{8.42} \\
\bottomrule
\end{tabular}
}
\hspace{0.015\linewidth}
\subcaptionbox{Impact of resolution-agnostic training.\label{tab:b}}[0.31\linewidth]{%
\centering\scriptsize
\begin{tabular}{lccc}
\toprule
\multirow{2}{*}{\textbf{Tokens}} & \multirow{2}{*}{\textbf{Generalist}} & \multicolumn{2}{c}{\textbf{gFID}} \\
\cmidrule(lr){3-4}
 &  & \textbf{w/o cfg} & \textbf{w/ cfg} \\
\midrule
\multirow{2}{*}{\textbf{64}}  & \no  & 33.15 & \textbf{8.42} \\
                              & \yes & \textbf{31.14} & 9.37 \\
\hdashline
\multirow{2}{*}{\textbf{128}} & \no  & \textbf{37.68} & \textbf{9.02} \\
                              & \yes & 39.50 & 10.58 \\
\bottomrule
\end{tabular}
}
\hspace{0.015\linewidth}
\subcaptionbox{Comparison with 
LlamaGen (fixed resolution).
\label{tab:c}}[0.35\linewidth]{%
\centering\scriptsize
\begin{tabular}{lcccc}
\toprule
\multirow{2}{*}{\textbf{Model}} & \multirow{2}{*}{\textbf{Tokens}} & \multirow{2}{*}{\textbf{Generalist}} & \multicolumn{2}{c}{\textbf{gFID}} \\
\cmidrule(lr){4-5}
 &  &  & \textbf{w/o cfg} & \textbf{w/ cfg} \\
\midrule
LlamaGen  & 576 & \no  & 33.44 & \textbf{7.15} \\
\rowcolor{hilite}
\tokenvibegen & \textbf{64} & \yes & \textbf{31.14} & 9.37 \\
\bottomrule
\end{tabular}
}

\caption{\textbf{Ablation study on ImageNet 256$\times$256 generations.} We perform a study on (a) the number of tokens, (b) generalist training, and (c) compare with baseline LlamaGen. All models are GPT-B variants and were trained for 100 epochs with a batch size of 256. 
}
\label{tab:gfid-ablations}
\end{table*}

\subsection{Image Reconstruction}

\paragraph{Experimental setup.}
We train two variants of \tokenvibe: \textbf{VibeToken-SL} (small encoder, large decoder) and \textbf{VibeToken-LL} (large encoder, large decoder).
Both are trained from scratch on ImageNet1k with mixed resolutions between $256{\times}256$ and $512{\times}512$; $60\%$ of the samples are square ($256^2$ or $512^2$) and the remainder are drawn from non-square aspect ratios.
We use a batch size of $64$ for $600\mathrm{k}$ iterations with cosine decay and peak learning rate $1\mathrm{e}{-4}$.
Unless noted otherwise, we train for dynamic latent lengths $L\in[32,256]$.
We adopt MVQ with $8$ codebooks of size $4096$ (effective vocabulary $32{,}768$).
All models are trained on a single node with $8{\times}$H100 GPUs; further hyperparameters are in the Appendix.

\paragraph{Evaluation protocol.}
We assess reconstruction on three different \textit{i.i.d.} and \textit{o.o.d.} regimes. 
We report rFID (lower is better) and vary $L$ where applicable.
\begin{enumerate}[nosep,leftmargin=*]
    \item ImageNet-1k (i.i.d.): $50\mathrm{k}$ validation images at $256^2$ and $512^2$.
    \item FFHQ (high-res): $10\mathrm{k}$ images at $1024^2$.
    \item Arbitrary aspect ratios (OOD): stress tests at aspect ratios $\{1:2,2:3,3:4,9:16,16:9,4:3,3:2,2:1\}$ with resolutions between $512^2$ and $1024^2$ using FFHQ.
\end{enumerate}

\begin{table*}[t]
\centering
\resizebox{0.8\textwidth}{!}{%
\begin{tabular}{l ccc ccc ccc}
\toprule
\multirow{2}{*}{\textbf{Model}} &
\multirow{2}{*}{\textbf{Type}} &
\multirow{2}{*}{\textbf{Params}} &
\multirow{1}{*}{\textbf{Arbitrary}} &
\multicolumn{3}{c}{\textbf{256$\times$256}} &
\multicolumn{3}{c}{\textbf{512$\times$512}} \\
\cmidrule(lr){5-7}\cmidrule(lr){8-10}
 &  &  & \textbf{Resolutions} & \textbf{Tokens} & \textbf{FID} & \textbf{IS} & \textbf{Tokens} & \textbf{FID} & \textbf{IS} \\
\midrule
MaskGIT~\cite{chang2022maskgit}              & \multirow{4}{*}{Mask.} & 177M & \no      & 256  & 4.02 & 355.6 & 1024 & 4.46 & 342   \\
TiTok-S/B-128~\cite{yu2024image} &      & 287M & \no      & 128  & 2.48 &  --   & 128    & 2.13 & --    \\
MAGVIT-v2~\cite{yu2023language}           &       & 307M & \no     & 256  & 1.78 & 319.4 & 1024 & 1.91 & 324.3 \\
MaskBit~\cite{weber2024maskbit}             &       & 305B & \no   & 256  & 1.52 & 328.6 &  --  &  --  &  --   \\
\midrule
VAR-d30~\cite{tian2024visual}             & \multirow{2}{*}{VAR}   & 2B   & \no     & 680  & 1.92 & 323.1 & 2240 & 2.63 & 303.2 \\
VAR-d30-re~\cite{tian2024visual}  &      & 2B   & \no    & 680  & 1.73 & 350.2 &  --  &  --  &  --   \\
\midrule
GPT2-re~\cite{esser2021taming}             & \multirow{6}{*}{AR}    & 1.4B & \no     &   256   & 5.20 & 280.3 &  --  &  --  &  --   \\
VIM-L-re~\cite{yu2021vector}            &       & 1.7B & \no    & 1024 & 3.04 & 227.4 &  --  &  --  &  --   \\
Open-MAGVIT2-XL~\cite{luo2024open}     &       & 1.5B & \no    & 256  & 2.33 & 271.77 & --  & --  & --    \\
LlamaGen-XXL~\cite{sun2024autoregressive}        &       & 1.4B & \no    & 576  & 2.34 & 253.91 & --  & --  & --    \\
UniTok-XXL~\cite{ma2025unitok}          &       & 1.4B & \no     & 256  & 2.51 & 216.7 &  --  &  --  &  --   \\
RAR-XXL~\cite{yu2025randomized}             &       & 1.4B & \no & 256 & 1.48 & 326.0 & 1024 & 1.66 & 295.7 \\
\hdashline\addlinespace[2pt]
\textsuperscript{{\color{BrickRed}{\textbf{Single AR model for all Resolutions}}}} &  \\[-3pt]

\tokenvibegen-B      & \multirow{2}{*}{AR}    & 87M & \yes     & \textbf{64}   & 7.62 & 185.92  & \textbf{64} & 7.48 & 189.43  \\
\tokenvibegen-XXL    &     & 1.5B & \yes      & \textbf{64}   & 3.62 & 226.93  & \textbf{64} & 3.60 & 230.33  \\
\bottomrule
\end{tabular}
}
\caption{\textbf{Comparison across models and resolutions.} Columns group 256$\times$256 and 512$\times$512 results (Tokens/FID/IS). 
Among many prior works on discrete image modeling, only \tokenvibegen\ scales to arbitrary resolutions effortlessly.
}
\label{tab:fixed-res-gfid-table}
\end{table*}

\begin{figure}[t]
    \centering
    \includegraphics[width=\linewidth]{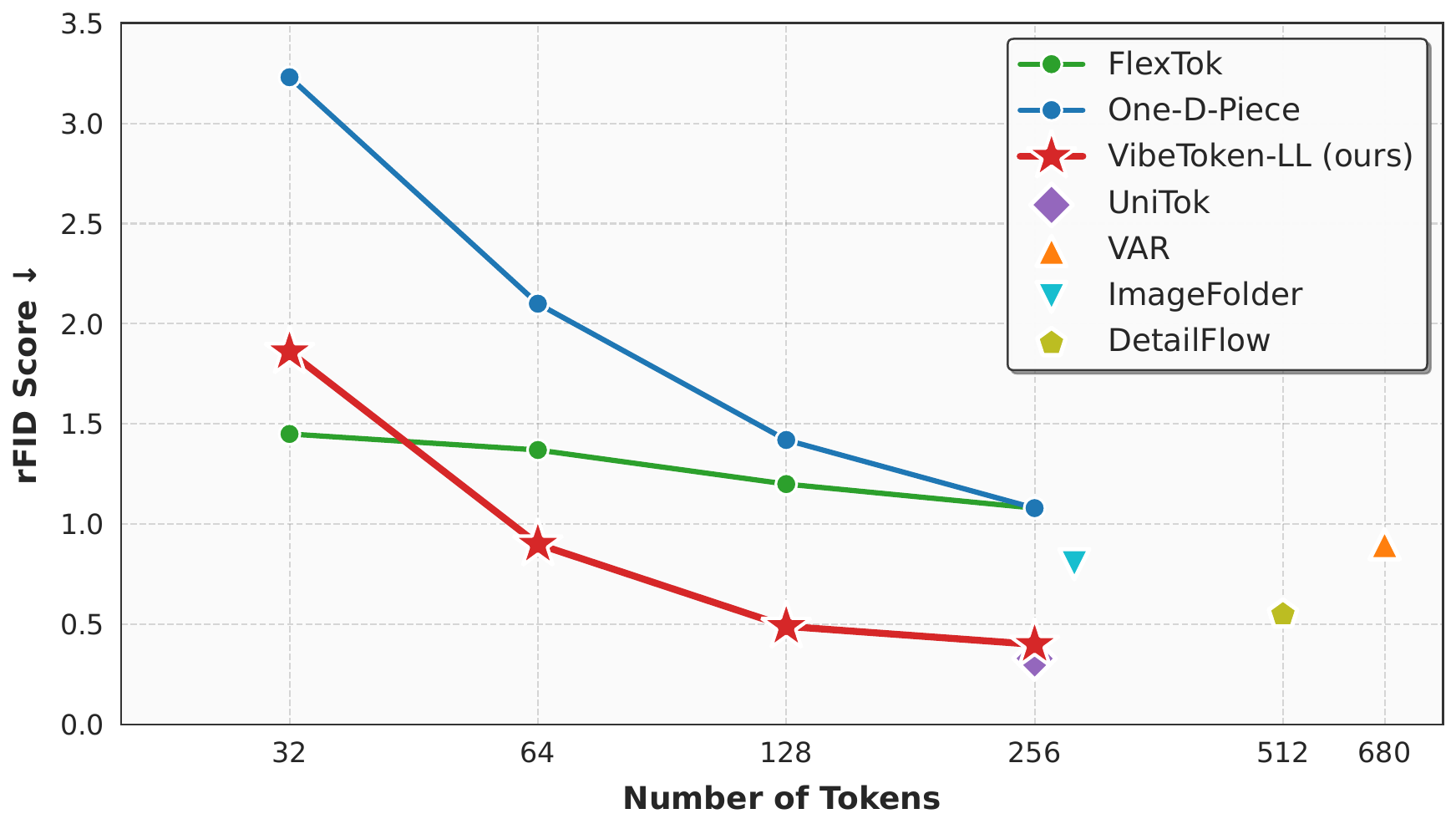}
    \caption{\textbf{Dynamic token length.} rFID vs.\ latent length $L$ at $256^2$. \tokenvibe\ is the only model that pairs competitive quality with \emph{resolution generalization}.
    }
    \label{fig:dynamic_length_recon}
\end{figure}

\paragraph{Main results.}
Table~\ref{tab:tokenizer-comparison} compares \tokenvibe\ with 2D and 1D tokenizers across resolutions.
Unlike prior 1D tokenizers (which generally assume fixed training grids), \tokenvibe\ \emph{natively} supports arbitrary resolutions and aspect ratios while retaining short, dynamic token sequences.
Concretely, \textbf{VibeToken-LL} attains \textbf{0.40} rFID at $256^2$, \textbf{0.51} at $512^2$, \textbf{2.40} at $1024^2$, and \textbf{3.60} on arbitrary-resolution stress tests; \textbf{VibeToken-SL} achieves $0.43/0.55/2.45/4.21$ on the same settings.
Among 1D baselines, UniTok reports a strong $0.33$ rFID at $256^2$ but does not support dynamic resolution; several other 1D methods lack results beyond $256^2$.
Against 2D tokenizers, \tokenvibe\ is competitive on $256^2$–$512^2$ (e.g., IBQ: $0.97/0.40$) while trading a modest performance drop at $1024^2$, \tokenvibe\ dramatically reduces token requirements at arbitrary high resolutions.
Figure~\ref{fig:dynamic_length_recon} shows the dynamic compression \textit{vs.} rFID performance trade-off.
With as few as \textbf{64 tokens}, \tokenvibe\ matches or surpasses prior dynamic-length 1D tokenizers at $256^2$; increasing to $128$–$256$ tokens yields further gains while preserving the ability to decode at arbitrary target resolutions.
We provide detailed ablations and native super-resolution performance in the appendix. 

\paragraph{Efficiency analysis.}
Figure~\ref{fig:tokenvibe-four-wide} plots tokenizer inference FLOPs across resolutions.
2D tokenizers scale roughly linearly with pixel count: e.g., IBQ grows from $\sim\!0.64\,$T to $\sim\!10.30\,$T FLOPs from 256$^2$ to 1024$^2$ resolution.
In contrast, \textbf{\tokenvibe\ maintains constant $\sim\!1.04\,$T FLOPs} (maximum), enabling a fixed compute budget independent of resolution.
This stability, combined with dynamic lengths ($L{=}$64–256), yields state-of-the-art throughput among flexible tokenizers.
Moreover, while 2D tokenizers have a fixed $16{\times}$/ $32{\times}$ compression, \tokenvibe\ achieves much higher effective compression as resolution increases: at $1024^2$, images can be encoded with \textbf{64 – 256 tokens}, improving token efficiency by \textbf{$16{\times}$ to $64{\times}$} relative to 2D counterparts; resulting into effective compression ratio of 64$\times$ to 128$\times$.

\subsection{Image Generation}

We evaluate the \emph{resolution-agnostic} tokens produced by \tokenvibe\ by training class-conditioned AR generators that can support arbitrary target resolutions. 
Because scaling prior AR models across resolutions with existing tokenizers is compute-intensive and impractical in academic settings, we compare against (i) open-source multi-resolution diffusion models and (ii) fixed-resolution AR baselines.

\paragraph{Experimental setup.}
We train \tokenvibegen\ on mixed resolutions from $256^2$ to $512^2$ within an ImageNet-1k compute envelope. We use Query–Key LayerNorm for stability, select \textbf{VibeToken-LL} as the default tokenizer, and keep the rest of the LlamaGen-style stack unchanged. Because \tokenvibe\ supplies small, resolution-agnostic sequences, the training compute remains comparable to (or lower than) a fixed $256{\times}256$ LlamaGen baseline. Additional details are in the Appendix.

\paragraph{Ablation study.}
Table~\ref{tab:gfid-ablations} reports GPT-B ablations over 100 epochs. 
We find that \textbf{64 tokens} are enough and yield the best gFID.
Notably, increasing the token length to 128/256 does not offer any gains.
This highlights that not all tokens important for improving rFID are needed to improve gFID.
Training a \emph{generalist} (mixed-resolution) model introduces a small degradation relative to single-resolution training, but the \textbf{64-token} setting remains competitive. Guided by these findings, we train \tokenvibegen-XXL generalist models with only 64 tokens and report the main results.

\paragraph{Main results.}
We benchmark class-conditional generation against both AR and diffusion families. Since prior AR models do not natively generalize across resolutions, our primary comparisons are to multi-resolution diffusion baselines. As shown in Table~\ref{tab:gfid-resolutions-combined}, \tokenvibegen\ \emph{smoothly} scales to higher and arbitrary resolutions and achieves near-SOTA gFID despite being trained within ImageNet-1k budgets. Concretely, it outperforms \textsc{EDM2} and \textsc{FiT-v2} across many resolution settings and is competitive with \textsc{NiT}. This demonstrates that AR models \emph{can} be made resolution-generalist on academic compute by delegating scale handling to a resolution-agnostic tokenizer. Finally, when evaluated strictly at $256^2$ or $512^2$ against models trained only at those resolutions (Table~\ref{tab:fixed-res-gfid-table}), \tokenvibegen\ can trail specialist baselines; we attribute this to the shared budget across resolutions during generalist training, which makes direct comparison to single-resolution specialists less informative.
We provide qualitative examples in the appendix.

\paragraph{Efficiency analysis.}
AR models are notoriously expensive to scale, especially when token counts grow with resolution. \tokenvibegen\ breaks this pattern by decoupling compute from $(H,W)$: with a fixed latent length $L$, inference cost is governed by $L$ and model depth—not by image resolution. As shown in Figure~\ref{fig:tokenvibe-four-wide}, \tokenvibegen-XXL requires \textbf{179\,G} \emph{per forward step} with $L{=}64$ for \emph{any} resolution,\footnote{Numbers reported per forward pass without KV caching; The total generation cost is therefore higher and scales with the sequence length $L$.} whereas a fixed-resolution LlamaGen scales up to $\sim\!11$\,T FLOPs to reach $1024^2$ even in idealized settings. In wall-clock terms, a diffusion SOTA baseline (NiT) takes $\mathbf{1.08}$\,s to synthesize a single $1024^2$ image at $\mathbf{5.87}$ gFID, while \textbf{\tokenvibegen-XXL} achieves \textbf{3.54} gFID with \textbf{0.46} s latency. 
These results underscore that resolution-agnostic tokens enable high-resolution AR generation with substantially lower and more predictable compute.

\section{Conclusion}

We present \tokenvibe\ and \tokenvibegen, a practical recipe for scaling autoregressive (AR) image generation to \emph{arbitrary} resolutions and aspect ratios under a fixed, user-controllable compute budget. 
The core idea is a \emph{resolution-agnostic} 1D tokenizer (\tokenvibe) that produces short, variable-length sequences (64-256 tokens) independent of the resolutions. 
This is achieved through dynamic grid positional embeddings, adaptive decoding, and length-aware training.
To our knowledge, \tokenvibe\ is the first 1D Transformer-based tokenizer to natively support arbitrary resolutions while retaining strong compression and reconstruction quality. Building on these tokens, \tokenvibegen\ achieves competitive class-conditional generation across resolutions, with inference FLOPs governed by token length rather than pixel count, thereby narrowing the gap between AR and diffusion pipelines within academic compute budget.

{
    \small
    \bibliographystyle{ieeenat_fullname}
    \bibliography{main}

}
\renewcommand{\thesection}{\Alph{section}}
\renewcommand{\thefigure}{\Alph{figure}}
\renewcommand{\thetable}{\Alph{table}}

\clearpage
\setcounter{page}{1}
\maketitlesupplementary

\section{Extended Related Works}

\paragraph{Image tokenizers.}
Tokenizers are central to modern image generation. Diffusion/flow models are typically built on continuous latents (\ie, VAEs) and achieve state-of-the-art results, whereas AR models commonly rely on discrete latents (\eg, VQ-VAE variants). Early AR systems use \emph{2D} tokenizers such as MAGViT~\cite{lee2022autoregressive}, Open-MAGViT2~\cite{luo2024open}, and LlamaGen~\cite{sun2024autoregressive}. A known limitation is that the token count grows with resolution—\eg, with stride $16$, a $1024{\times}1024$ image yields $4096$ tokens—making scaling costly. DA-AR reduces tokens via stronger downsampling (e.g., $32{\times}$ compression $\Rightarrow$ $1024$ tokens at $1024^2$), and CAT introduces LLM-guided \emph{dynamic} compression for 2D VAEs; however, these encodings remain only partially flexible and still couple token length to spatial resolution.
Recent work proposes \emph{1D} tokenizers with higher compression. TiTok~\cite{yu2024image} introduced the paradigm; TA-TiTok~\cite{kim2025democratizing} extends training (with/without text conditioning). Subsequent approaches refine objectives, quantization, and training curricula~\cite{Chen2025MaskedAA, chen2025softvq, Qiu2025ImageTN}. Despite strong compression, many early 1D designs assume fixed training grids and output a \emph{fixed} latent length, limiting generalization to unseen resolutions/aspect ratios.
One-D-Piece and FlexTok~\cite{miwa2025one,bachmann2025flextok} enable \emph{dynamic} latent lengths (e.g., $1$–$256$ tokens) but primarily at fixed resolutions, so cross-resolution generalization remains challenging. Instella-T2I~\cite{wang2025instella} and Layton~\cite{Xie2025LaytonLC} scale 1D tokenizers to $1024^2$ using as few as $128$–$256$ tokens, yet the encodings are still tied to specific canvases and do not natively handle arbitrary aspect ratios. A-Token~\cite{Lu2025ATokenAU} moves toward arbitrary resolutions, but requires careful high-resolution training and remains sensitive to data/compute budgets.
In contrast, \tokenvibe\ scales 1D tokenization to \emph{resolution-agnostic} encodings: it decouples latent length from spatial size, supports dynamic token budgets, and generalizes across resolutions and aspect ratios without retraining per resolution. This flexibility enables the AR generator to operate with a fixed, user-controllable compute budget while retaining strong fidelity.

\paragraph{Dynamic Resolution-adaptation for Transformers.}
Vision Transformers (ViT)~\cite{dosovitskiy2020image} are typically trained on fixed patch grids, making \emph{length generalization} difficult due to absolute positional encodings and quadratic attention costs. A line of work improves flexibility primarily for \emph{discriminative} tasks: ResFormer~\cite{tian2023resformer}, DeiT~\cite{touvron2021training}, ViTAR~\cite{fan2024vitar}, and NaViT~\cite{dehghani2023patch} explore position encodings, packed batching, and attention variants to better handle variable input sizes, with ViTAR/NaViT showing notably stronger scaling than their baselines. FlexiViT~\cite{beyer2023flexivit} enables \emph{compute-adaptive} inference via dynamic patch sizes. Beyond absolute PEs, recent methods leverage Axial-RoPE and learned RoPE variants to improve cross-resolution extrapolation~\cite{heo2024rotary}. On the \emph{generative} side, Diffusion Transformers (DiTs)~\cite{peebles2023scalable} have been scaled to native resolutions with FiT-v1/v2~\cite{lu2024fit, wang2024fitv2} and NiT~\cite{wang2025native} through RoPE-based encodings and training strategies inspired by NaViT. However, autoregressive (AR) image generation remains largely fixed-resolution: token counts grow with resolutions and next-token training is brittle to length shifts.
We bridge this gap by pairing a \emph{resolution-agnostic} 1D tokenizer, \tokenvibe, with a LlamaGen-style AR head, \tokenvibegen. The tokenizer decouples token \emph{length} from spatial resolution, while the generator conditions on the target canvas; together, they enable arbitrary resolutions and aspect ratios under a constant, user-controllable token budget.

\section{Experimental Setup}
This section details the training and inference setups for \tokenvibe\ (tokenizer) and \tokenvibegen\ (AR generator).
\begin{table}[ht]
\centering
\renewcommand{\arraystretch}{1.15}
\setlength{\tabcolsep}{6pt}
\scriptsize
\resizebox{\linewidth}{!}{
\begin{tabular}{
    l
    >{\centering\arraybackslash}p{2.6cm}
    >{\centering\arraybackslash}p{2.6cm}
}
\toprule
\textbf{Config} & \textbf{Ablations} & \textbf{VibeToken} \\
\midrule
Model Type                       & VAE          & MVQ \\
Token size                 & 16           & 256 \\
\# of codebooks            & ---          & 8 \\
Vocab size                 & ---          & 32768 \\
Encoder/Decoder            & SS           & SL/LL \\
Discriminator start        & 100{,}000    & 300{,}000 \\
Quantizer weight           & 1            & 1 \\
Discriminator weight       & 0.1          & 1 \\
Perceptual weight          & 1.1          & 1.1 \\
Reconstruction weight      & 1            & 1 \\
Commitment cost            & ---          & 0.25 \\
KL weight                  & $1\times10^{-6}$ & --- \\
\# of tokens               & 256          & 32--256 \\
\midrule
Optimizer                  & \multicolumn{2}{c}{AdamW} \\
Learning rate              & \multicolumn{2}{c}{0.0001} \\
GAN LR                     & \multicolumn{2}{c}{0.0001} \\
$\beta_1$                  & \multicolumn{2}{c}{0.9} \\
$\beta_2$                  & \multicolumn{2}{c}{0.999} \\
Weight decay               & \multicolumn{2}{c}{0.0001} \\
Scheduler                  & \multicolumn{2}{c}{Cosine} \\
Warmup steps               & \multicolumn{2}{c}{10{,}000} \\
\midrule
Dataset                    & ImageNet1k   & ImageNet1k \\
Batch size / GPU           & 32           & 8 \\
Gradient accumulation      & 1            & 2 \\
GPUs                       & 4$\times$H100s & 8$\times$H100s \\
Precision                  & bf16         & bf16 \\
Training steps             & 200{,}000    & 600{,}000 \\
Variable resolutions       & \multicolumn{1}{p{2.6cm}}{\raggedright\texttt{\{"256x256": 0.5, "512x512": 0.5\}}} &
                             \multicolumn{1}{p{2.6cm}}{\raggedright\texttt{\{"256x256": 0.3, "512x512": 0.3, "384x256": 0.1, "256x384": 0.1, "512x384": 0.1, "384x512": 0.1\}}} \\
\bottomrule
\end{tabular}
}
\caption{\tokenvibe\ pretraining setup.}
\label{tab:vibetoken_hyperparams}

\end{table}

\begin{table}[t]
\centering
\renewcommand{\arraystretch}{1.15}
\setlength{\tabcolsep}{6pt}
\scriptsize
\resizebox{\linewidth}{!}{
\begin{tabular}{lcc}
\toprule
\textbf{Config} & \textbf{GPT-B} & \textbf{GPT-XXL} \\
\midrule
\multicolumn{3}{l}{\textbf{Tokenizer \& Data}} \\
Tokenizer                & \multicolumn{2}{c}{VibeToken-MVQ-LL} \\
Tokens                   & \multicolumn{2}{c}{64 / 128 / 256} \\
Resolutions              & \multicolumn{2}{p{6cm}}{\raggedright\texttt{\{"256x256": 0.3, "512x512": 0.3, "384x256": 0.07, "256x384": 0.07, "512x384": 0.07, "384x512": 0.07, "256x512": 0.06, "512x256": 0.06\}}} \\
Encoder patch size        & \multicolumn{2}{c}{16x16 (fixed for simplicity)} \\
Dataset                  & \multicolumn{2}{c}{ImageNet1k} \\
\midrule
\multicolumn{3}{l}{\textbf{Model Architecture}} \\
Prediction head layers    & \multicolumn{2}{c}{4} \\
\# of codebooks           & \multicolumn{2}{c}{8} \\
Vocab size                & \multicolumn{2}{c}{32768} \\
Sequence length           & \multicolumn{2}{c}{64 / 128 / 256} \\
Class-dropout prob.       & \multicolumn{2}{c}{0.1} \\
Additional norm layers    & \multicolumn{2}{c}{QK} \\
Dropout prob.             & \multicolumn{2}{c}{0.1} \\
\midrule
\multicolumn{3}{l}{\textbf{Training Setup}} \\
Epochs                    & 300 & 150 \\
Optimizer                 & \multicolumn{2}{c}{AdamW} \\
Learning rate             & \multicolumn{2}{c}{0.0001} \\
$\beta_1$                 & \multicolumn{2}{c}{0.9} \\
$\beta_2$                 & \multicolumn{2}{c}{0.95} \\
Weight decay              & \multicolumn{2}{c}{0.05} \\
Precision                 & \multicolumn{2}{c}{None} \\
Schedule                  & \multicolumn{2}{c}{Linear} \\
GPUs                      & 4$\times$H100--8$\times$H100 & 8$\times$H100 \\
\bottomrule
\end{tabular}
}
\caption{Hyperparameters for \textit{GPT-B} and \textit{GPT-XXL} configurations of \tokenvibegen.}
\label{tab:gpt-configs}

\end{table}

\subsection{VibeToken}
Table~\ref{tab:vibetoken_hyperparams} summarizes the hyperparameters for \tokenvibe\ and its ablations. We follow TiTok/TA-TiTok conventions, adopting TA-TiTok's single-stage training (no text conditioning).

\paragraph{Ablations.}
We train a small encoder/decoder VAE for 200{,}000 iterations on $4{\times}$H100 with per-GPU batch size $32$. The latter half of training uses an adversarial loss as well. Ablation models are trained at two resolutions, $256{\times}256$ and $512{\times}512$.

\paragraph{Final tokenizer.}
The scaled \tokenvibe\ uses multi-vector quantization (MVQ) with $8$ codebooks and $256$ latent dimensions per token, factorized into $8$ sub-codes of $32$ dimensions each. The maximum latent length is variable in $[32,256]$; each token comprises $8$ sub-tokens \emph{without} increasing the AR sequence length. Concretely, following the RQ-Transformer style prediction-head, the AR embedding/output layers combine the $8$ sub-codes within the channel dimension after UniTok, so time length $L$ is unchanged.

\paragraph{Training schedule.}
We train the \textbf{SL} and \textbf{LL} variants for 600{,}000 iterations on a single node with $8{\times}$H100 GPUs, batch size $8$ per GPU, and gradient accumulation $2$ (effective batch $128$). Training images are sampled over six resolutions/aspect ratios between $256{\times}256$ and $512{\times}512$ with probabilities listed in Table~\ref{tab:vibetoken_hyperparams}. Despite training below $512^2$, \tokenvibe\ generalizes to higher resolutions ($1024^2$) without sacrificing reconstruction performance.

\subsection{VibeToken-Gen}
Table~\ref{tab:gpt-configs} lists pretraining settings for \tokenvibegen\ \textbf{B} ($\approx 90$M) and \textbf{XXL} ($\approx 1.4$B). Aside from model size, the setups are identical where possible (epochs and GPU count scale with size).

\paragraph{Tokenizer and tokens.}
We fix the tokenizer to \tokenvibe\texttt{-MVQ-LL}. Tokens of length $L\in\{64,128,256\}$ are extracted over eight resolutions according to Table~\ref{tab:gpt-configs}. For simplicity, the encoder patch size is $16{\times}16$.

\paragraph{Model and training.}
Following LlamaGen, we train on ImageNet-1k with ten-crop augmentation. We apply QK LayerNorm for stability. To predict MVQ sub-codes efficiently, we attach a lightweight 4-layer residual transformer head after UniTok that predicts the $8$ sub-codes per token; this reduces FLOPs by keeping the temporal length $L$ fixed while factorizing predictions across sub-codes. We observed instability with \texttt{bfloat16}; all AR training is therefore conducted in \texttt{fp32}.

\paragraph{Inference and evaluation.}
Unless stated otherwise, quantitative results use classifier-free guidance (CFG) fixed across runs with sampling parameters: temperature$=1.0$, top-$k{=}0$, top-$p{=}1.0$. Qualitative samples use CFG$=4.0$, temperature$=0.9$, top-$k{=}500$, top-$p{=}1.0$. The decoder patch size is fixed to $16{\times}16$ for resolutions $\le 512{\times}512$ and $32{\times}32$ for higher resolutions.

\section{Ablations}

\subsection{VibeToken}

\begin{table}[t]
\centering
\resizebox{\linewidth}{!}{
\begin{tabular}{l|ccc|ccc}
\toprule
\textbf{Model} & \textbf{Type} & \textbf{NFEs} & \textbf{Scale} &
\textbf{PSNR$\uparrow$} & \textbf{SSIM$\uparrow$} & \textbf{LPIPS$\downarrow$} \\
\midrule
SDXL-Upscaler           & Diffusion & 20 & \multirow{3}{*}{2$\times$} & \textbf{27.97} & 0.710 & 0.317 \\
Flux-Upscaler           & Diffusion & 20 &  & 25.17          & 0.609 & 0.377 \\
\rowcolor{hilite}
\tokenvibe-LL   & VQ-VAE    &  1 &  & 24.98          & \textbf{0.838} & \textbf{0.261} \\
\midrule
SDXL-Upscaler           & Diffusion & 20 & \multirow{3}{*}{4$\times$} & \textbf{29.10} & 0.721 & 0.361 \\
Flux-Upscaler           & Diffusion & 20 &  & 25.88          & 0.637 & 0.357 \\
\rowcolor{hilite}
\tokenvibe-LL   & VQ-VAE    &  1 &  & 24.11          & \textbf{0.805} & \textbf{0.310} \\
\midrule
\midrule
\tokenvibe-LL & VQ-VAE & 1 & 1$\times$ & 24.91 & 0.839 & 0.263 \\
\bottomrule
\end{tabular}
}
\caption{\textbf{Super-resolution ablation.}
Ground truth high resolution is fixed to 1024$\times$1024 and bicubic downsampling is used to get low resolution counterparts.}
\label{tab:sr-recon}
\end{table}

\begin{figure*}[t]
    \centering
    \includegraphics[width=\linewidth]{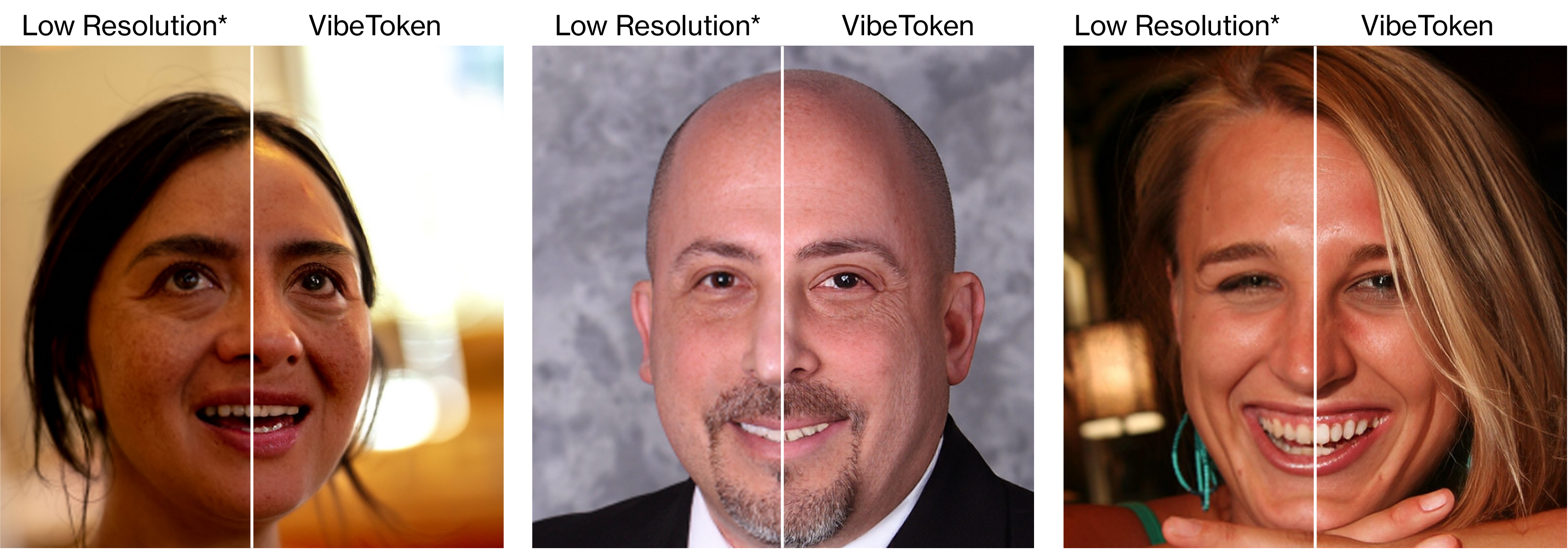}
    \caption{\textbf{Super-resolution Task.} Qualitative examples of native 4$\times$ super-resolution ability of \tokenvibe\ with respect to simple bilinear resize operation.
    }
    \label{fig:sr_qualitative}
\end{figure*}

\paragraph{Image Super-Resolution.}
Table~\ref{tab:sr-recon} evaluates the \emph{native} super-resolution capability of \tokenvibe\ against diffusion/flow upsamplers (e.g., SDXL/Flux upsamplers). Existing tokenizers generally \emph{lack} native SR and require a separately trained upsampler; \tokenvibe\ does not. We use $10\text{k}$ FFHQ images at $1024{\times}1024$, bicubically downsample to $256{\times}256$ and $512{\times}512$ for $4{\times}$ and $2{\times}$ SR, respectively, and report PSNR/SSIM/LPIPS. Diffusion upsamplers tend to achieve higher PSNR (favoring smoothness), while \tokenvibe\ yields stronger SSIM and lower LPIPS, indicating better structure and perceptual fidelity. Thus, \tokenvibe\ delivers competitive SR \emph{without} high-resolution pretraining or an extra upsampler. Qualitative examples are shown in Figure~\ref{fig:sr_qualitative}.

\begin{figure}[t]
    \centering
    \includegraphics[width=\linewidth]{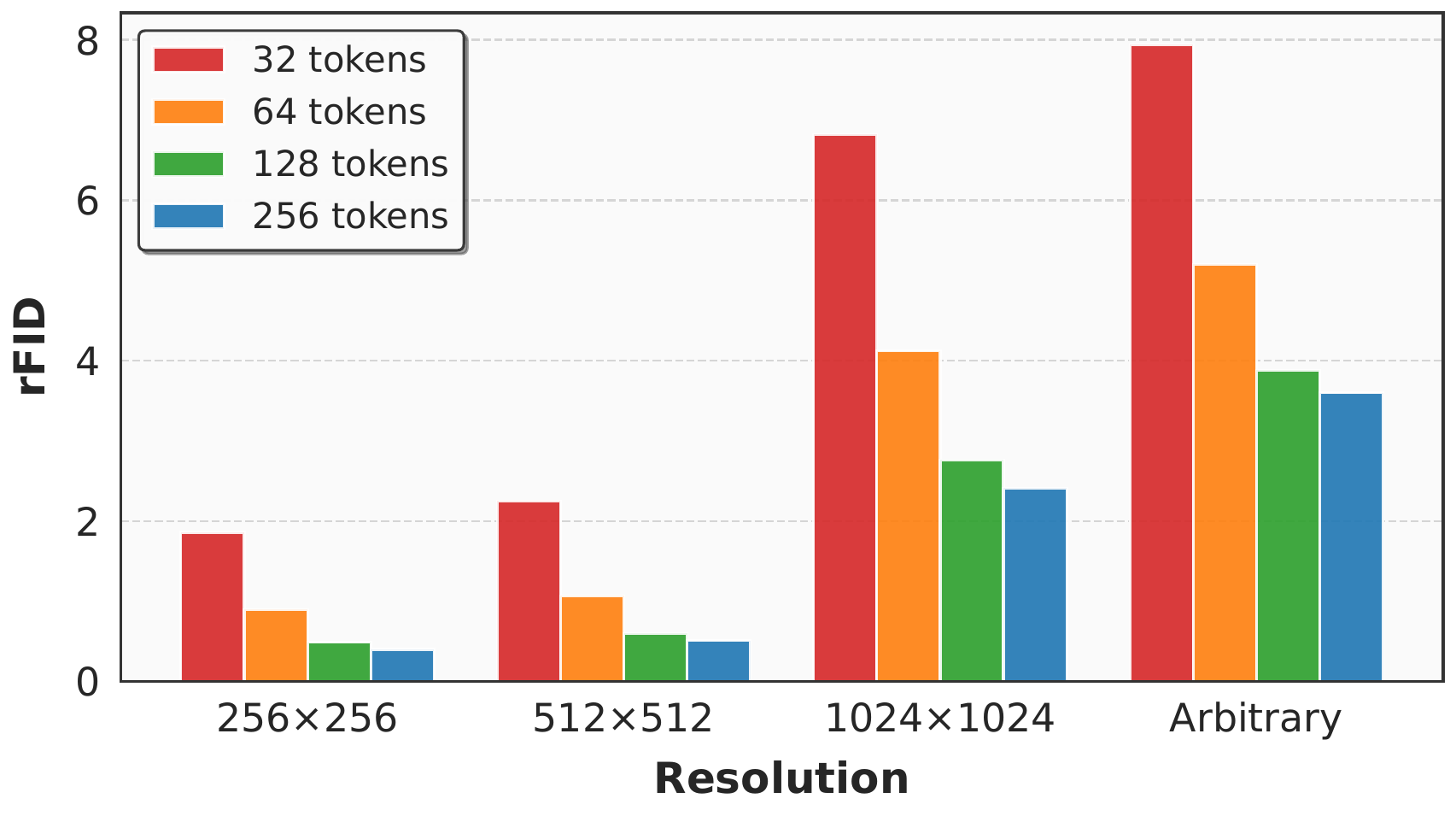}
    \caption{\textbf{Token-length Analysis.} rFID performance of \tokenvibe-LL with different token lengths across the resolutions. Notably, reference images are different for each resolutions, hence, rFID scale and performance is not comparable between each resolutions.
    }
    \label{fig:recon_tokens_vs_resolutions}
\end{figure}

\paragraph{Token Length \textit{vs.} Resolutions.} 
Figure~\ref{fig:recon_tokens_vs_resolutions} shows that \tokenvibe\ maintains strong reconstruction quality across resolutions while supporting dynamic-length encoding. In practice, \textbf{128 tokens} offer an excellent quality–compute balance for reconstruction; \textbf{256 tokens} provide small additional gains at higher cost. Notably, for \emph{generation}, \textbf{64 tokens} are most effective—later tokens primarily capture high-frequency details that benefit reconstruction more than synthesis. This separation suggests using short sequences for AR generation and longer sequences when reconstruction fidelity is paramount.


\begin{table*}[t]
\centering
\scriptsize
\setlength{\tabcolsep}{4pt}
\renewcommand{\arraystretch}{1.12}
\resizebox{\textwidth}{!}{
\begin{tabular}{l c | cccc | cccc}
\toprule
\multirow{2}{*}{Model} & \multirow{2}{*}{Type} &
\multicolumn{4}{c|}{\textbf{ImageNet 256$\times$256}} &
\multicolumn{4}{c}{\textbf{ImageNet 512$\times$512}} \\
\cmidrule(lr){3-6}\cmidrule(lr){7-10}
& &
\textbf{PSNR$\uparrow$} & \textbf{SSIM$\uparrow$} & \textbf{LPIPS$\downarrow$} & \textbf{rFID$\downarrow$} &
\textbf{PSNR$\uparrow$} & \textbf{SSIM$\uparrow$} & \textbf{LPIPS$\downarrow$} & \textbf{rFID$\downarrow$} \\
\midrule
MAGVIT-v2    & 2D & 22.62 & 0.5870 & 0.2050 & 1.17 & \underline{24.67} & 0.6510 & 0.2090 & \underline{0.50} \\
IBQ          & 2D & 22.95 & 0.5960 & 0.1980 & 0.97 & \textbf{25.19} & 0.6620 & 0.2010 & \textbf{0.40} \\
LLaMAGen     & 2D & 21.44 & 0.5320 & 0.2240 & 2.10 & 23.76 & 0.6140 & 0.2180 & 0.65 \\
\midrule
TiTok-B      & 1D & - & - & - & 1.70 & - & - & - & 1.52 \\
VAR          & 1D & - & - & - & 0.90 & - & - & - & - \\
ImageFolder   & 1D & - & - & - & 0.80 & - & - & - & - \\
UniTok       & 1D & - & - & - & 0.33 & - & - & - & - \\
One-D-Piece-L        & 1D & 19.04 & - & - & 1.08 & - & - & - & - \\
DetailFlow   & 1D & - & - & - & 0.55 & - & - & - & - \\
Instella     & 1D & - & - & - & - & 22.25 & 0.7040 & - & 1.32 \\
\hdashline
\tokenvibe-SL  & 1D & \underline{24.47} & \underline{0.8069} & \underline{0.1137} & \underline{0.43} & 22.86 & \underline{0.7541} & \underline{0.1998} & 0.55 \\
\textbf{\tokenvibe-LL} & 1D & \textbf{25.04} & \textbf{0.8194} & \textbf{0.1048} & \textbf{0.40} &
23.37 & \textbf{0.7649} & \textbf{0.1867} & 0.51 \\
\bottomrule
\end{tabular}
}
\caption{ImageNet reconstruction results across the diverse metrics at 256$\times$256 and 512$\times$512. Best is shown in bold and second best is underlined. \tokenvibe\ consistently outperforms 1D tokenizers and performs competitively to 2D tokenizer while having higher compression rates.}
\label{tab:imagenet_extended}

\end{table*}

\begin{table}[t]
\centering
\resizebox{\linewidth}{!}{
\begin{tabular}{l|ccc|ccc}
\toprule
& \multicolumn{3}{c|}{\textbf{256$\times$256}} & \multicolumn{3}{c}{\textbf{512$\times$512}} \\
\textbf{Model} & \textbf{FID$\downarrow$} & \textbf{PSNR$\uparrow$} & \textbf{SSIM$\uparrow$} &
\textbf{FID$\downarrow$} & \textbf{PSNR$\uparrow$} & \textbf{SSIM$\uparrow$} \\
\midrule
IBQ (2D)               & 6.14 & 22.56 & 0.603 & \textbf{4.12} & \textbf{24.50} & 0.656 \\
TiTok-SL/256           & 6.84 & 21.60 & 0.726 & -- & -- & -- \\
One-D-Piece-LL/256     & 8.02 & 18.41 & 0.612 & -- & -- & -- \\
\rowcolor{hilite}
\tokenvibe-LL/256 (ours)& \textbf{3.59} & \textbf{24.71} & \textbf{0.828} & \underline{4.46} & \underline{22.68} & \textbf{0.757} \\
\bottomrule
\end{tabular}
}
\caption{Comparison of \tokenvibe-LL on MSCOCO validation set across $256\times256$ and $512\times512$ resolutions. Best results are in bold, and second-best are underlined.}
\label{tab:mscoco_ablation}
\end{table}

\paragraph{Generalization beyond ImageNet.}
Table~\ref{tab:mscoco_ablation} compares the \tokenvibe-LL performance on MSCOCO for two resolutions along with IBQ and other 1D tokenizers. 
It can be observed that despite \tokenvibe~only trained on ImageNet (single entity focused) it generalizes to more complex images and achieves SoTA performance on 256$\times$256 resolution and gets competitive performance to 2D tokenizer.

\paragraph{Extended Reconstruction Results.}
Table~\ref{tab:imagenet_extended} provides the detailed results on ImageNet across reconstruction metrics such as PSNR, SSIM, LPIPS and rFID. 
\tokenvibe~\ consistently outperforms the 1D tokenizers and performs competitively to 2D tokenizers.
Importantly, 2D tokenizers have 1024 tokens for 512$\times$512 resolutions while \tokenvibe\ only requires 256 tokens and can support as low as 64 tokens.  

\subsection{VibeToken-Gen}

\begin{figure}[t]
    \centering
    \includegraphics[width=\linewidth]{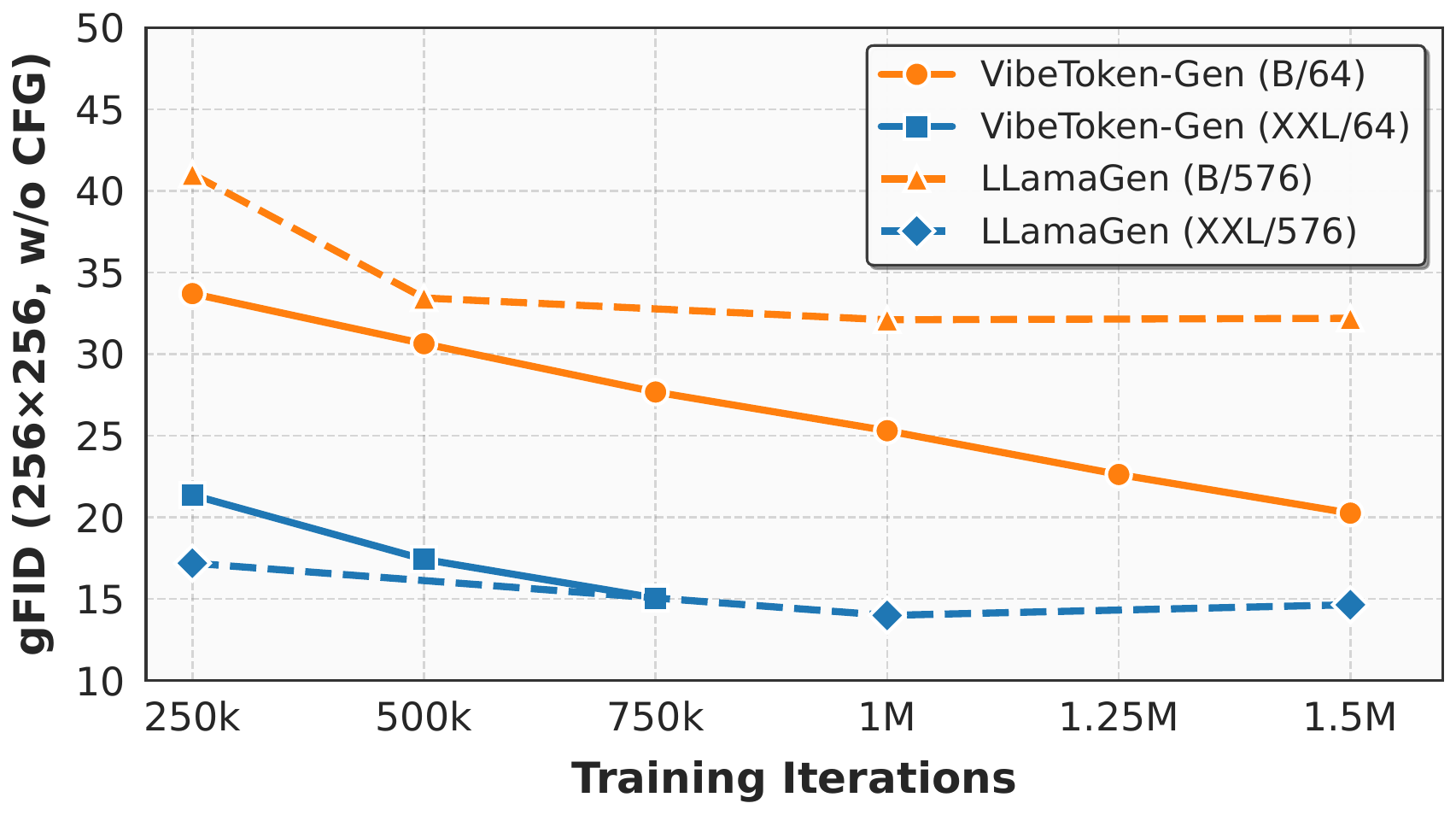}
    \caption{\textbf{Convergence Analysis.} gFID performance of \tokenvibegen\ and baseline LlamaGen models as training progresses. The results are shown on 256$\times$256 without classifier-free guidance.
    }
    \label{fig:convergence_vs_size}
\end{figure}

\paragraph{Convergence.} 
Figure~\ref{fig:convergence_vs_size} compares training dynamics of \tokenvibegen\ and LlamaGen at Base and XXL variants, evaluated on $256^2$ \emph{without} CFG to expose true likelihood learning. \tokenvibegen\ continues improving with training and surpasses LlamaGen, whereas LlamaGen saturates early and shows limited gains thereafter. This indicates better scaling of \tokenvibegen, attributable to shorter sequences and resolution-agnostic 1D tokens. Due to compute limits, we train \tokenvibegen\texttt{-XXL/64} to $750\text{k}$ iterations ($\approx\!150$ epochs); based on the \texttt{B/64} trend, we expect further improvements with longer training.

\begin{figure}[t]
    \centering
    \includegraphics[width=\linewidth]{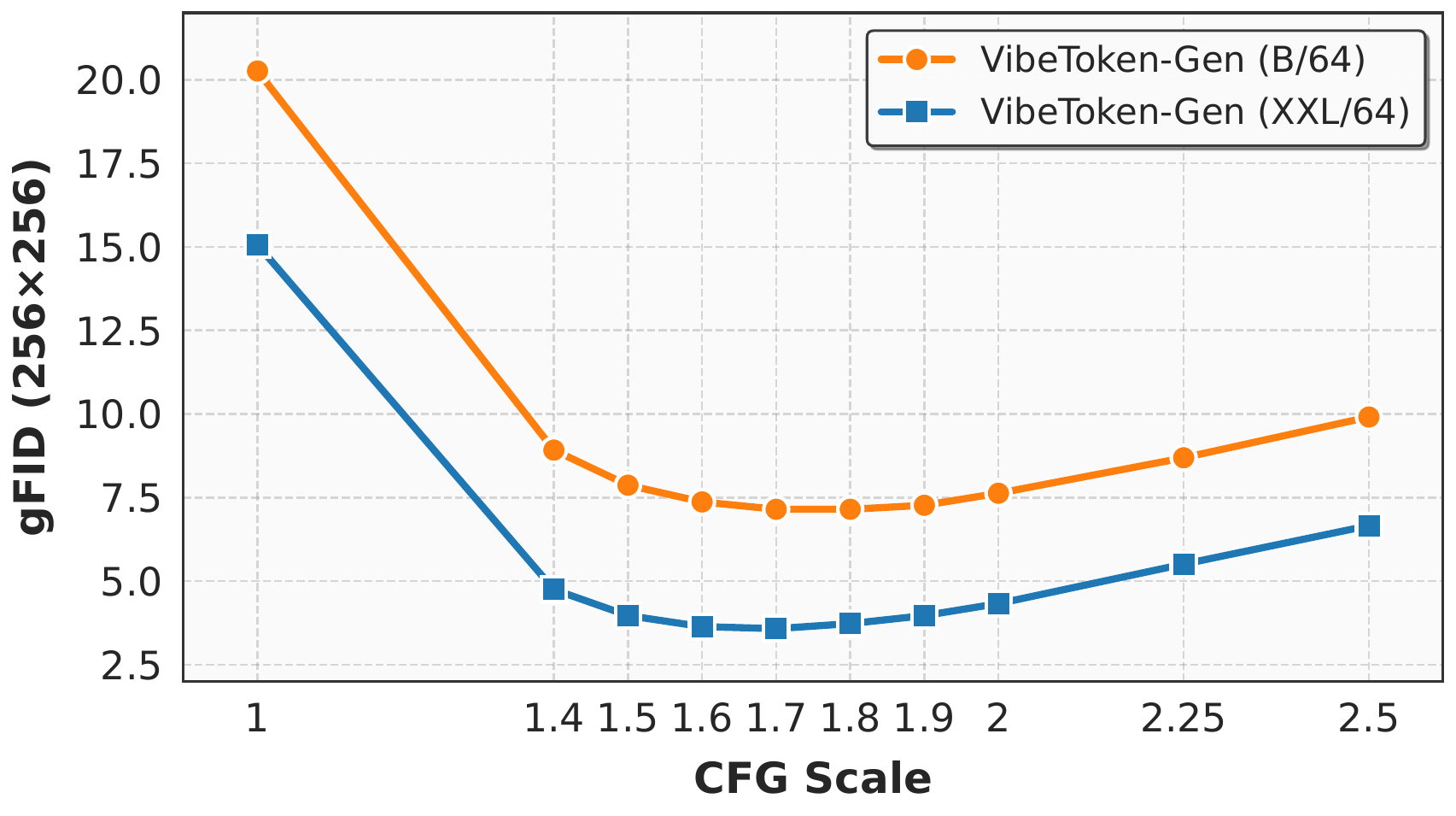}
    \caption{\textbf{CFG Analysis.} gFID performance of \tokenvibegen-B/XXL with respect to classifier-free guidance scale on 256$\times$256.
    }
    \label{fig:cfg_scale}
\end{figure}

\paragraph{Impact of classifier-free guidance.}
Figure~\ref{fig:cfg_scale} shows smooth behavior across CFG scales. After convergence, both \texttt{B} and \texttt{XXL} models achieve their best gFID around \mbox{$1.7$–$1.8$}. Accordingly, we use \mbox{CFG$=1.75$} and \mbox{CFG$=2.0$} when reporting \tokenvibegen\ results.


\begin{table}[t]
\centering
\begin{tabular}{lcc}
\toprule
\textbf{Model vs. Sec/Img} & \textbf{256$\times$256} & \textbf{1024$\times$1024} \\
\midrule
\multicolumn{3}{l}{\textit{Tokenizer enc./dec. only}} \\
LlamaGen-Tok      & \textbf{0.005} & 0.082 \\
\rowcolor{hilite}
\tokenvibe-LL      & 0.017          & \textbf{0.017} \\
\midrule
\multicolumn{3}{l}{\textit{End-to-end incl. tok. dec. + generation}} \\
LlamaGen / XXL    & \textbf{0.20} & 32.79 \\
\rowcolor{hilite}
\tokenvibegen\ / XXL & 0.46         & \textbf{0.46} \\
\bottomrule
\end{tabular}
\caption{Inference speed (seconds per image) comparison across resolutions.}
\label{tab:speed}
\end{table}

\paragraph{Efficiency Analysis.}
Table~\ref{tab:speed} shows the comparison of \tokenvibe\ and \tokenvibegen\ for reconstruction and generation tasks with respect to LlamaGen.
It can be observed that LlamaGen (theoretically) requires 32 seconds to generate 1024$\times$1024 resolution images while \tokenvibegen\ can do sub-one second generations.

\section{Qualitative Results}

\begin{figure*}[t]
    \centering
    \includegraphics[width=\linewidth]{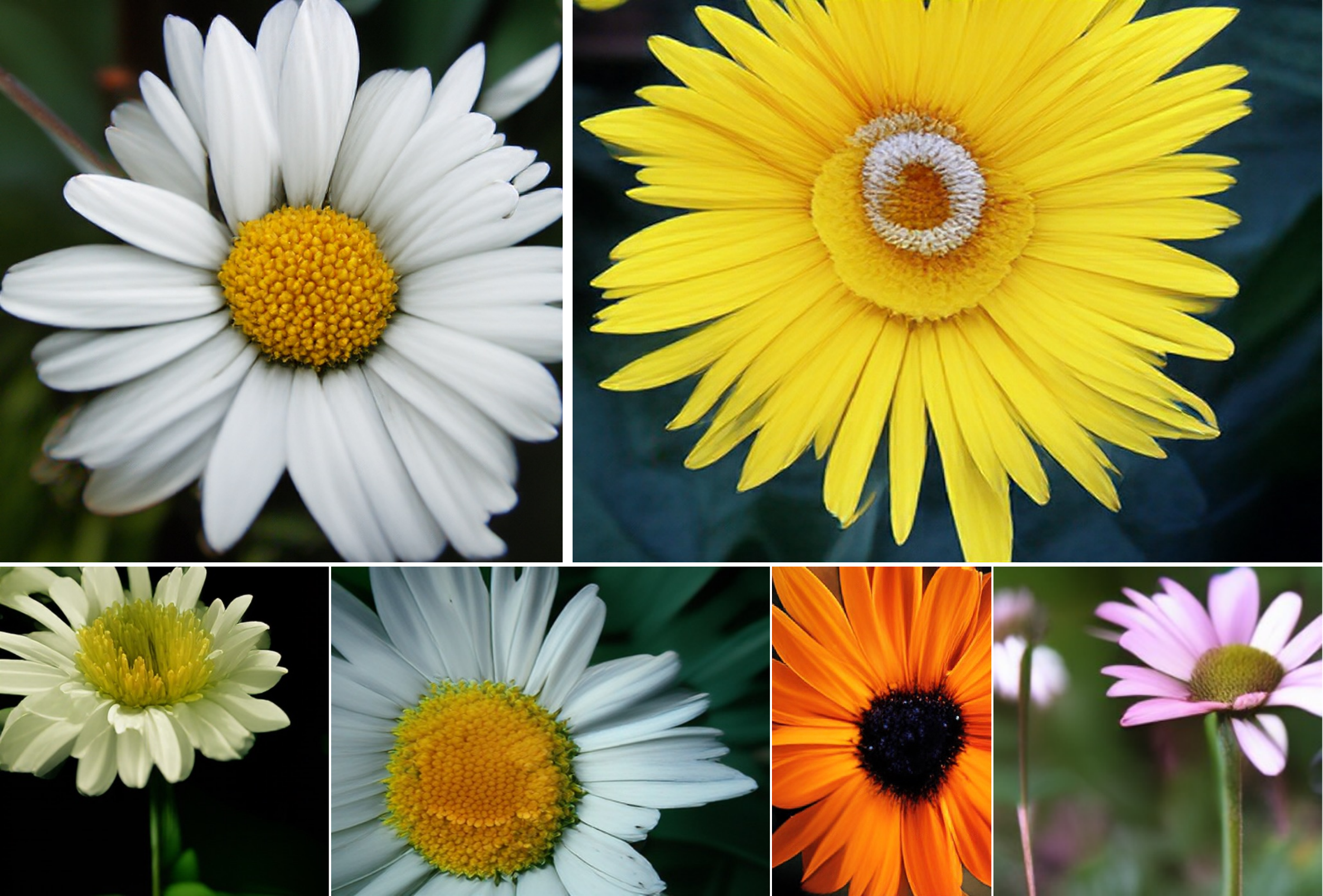}
    \caption{\textbf{Class 985.} Qualitative examples of generated images using \tokenvibegen\ (XXL/64) on arbitrary resolutions.
    }
    \label{fig:class_985}
\end{figure*}

\begin{figure*}[t]
    \centering
    \includegraphics[width=\linewidth]{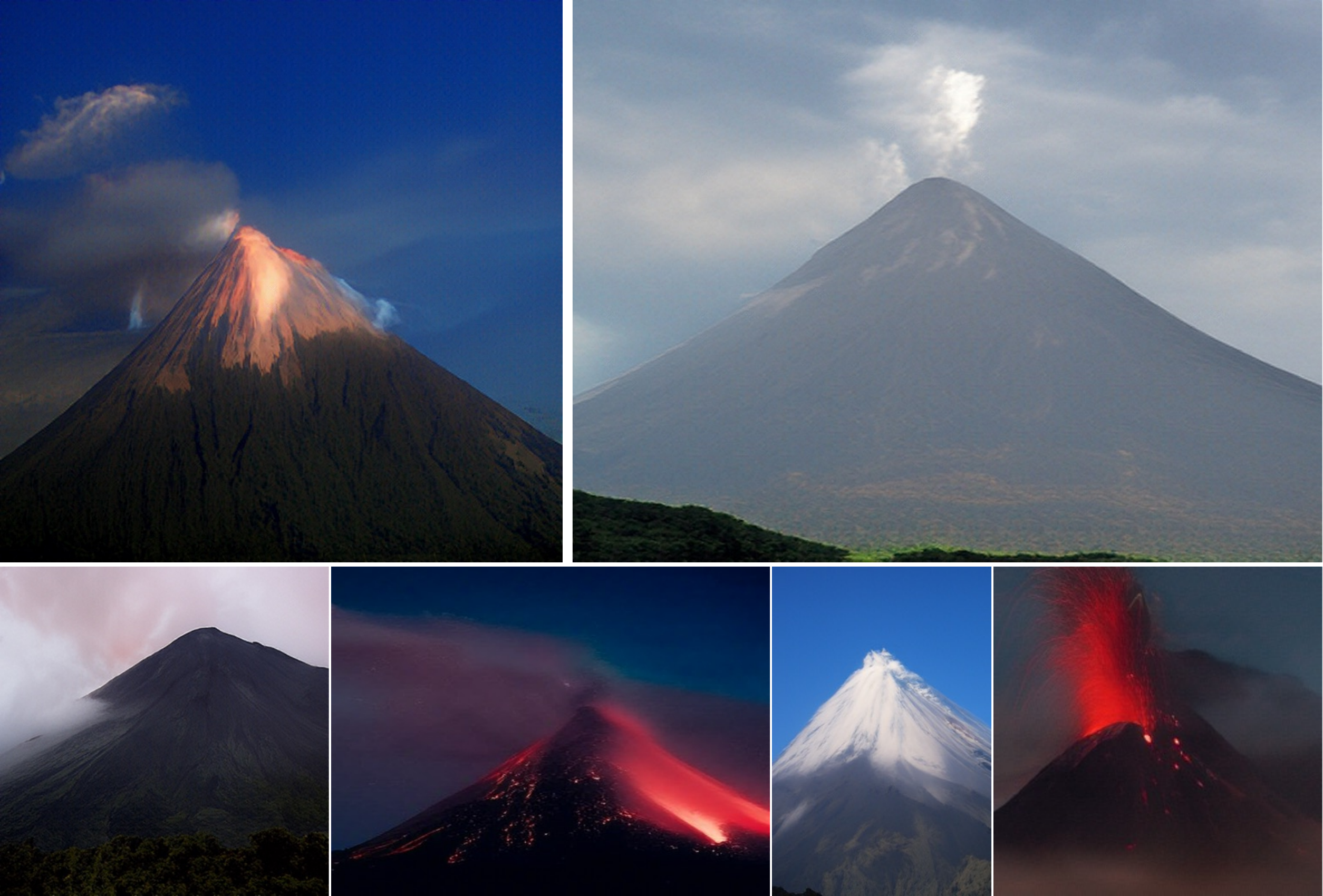}
    \caption{\textbf{Class 980.} Qualitative examples of generated images using \tokenvibegen\ (XXL/64) on arbitrary resolutions.
    }
    \label{fig:class_980}
\end{figure*}

\begin{figure*}[t]
    \centering
    \includegraphics[width=\linewidth]{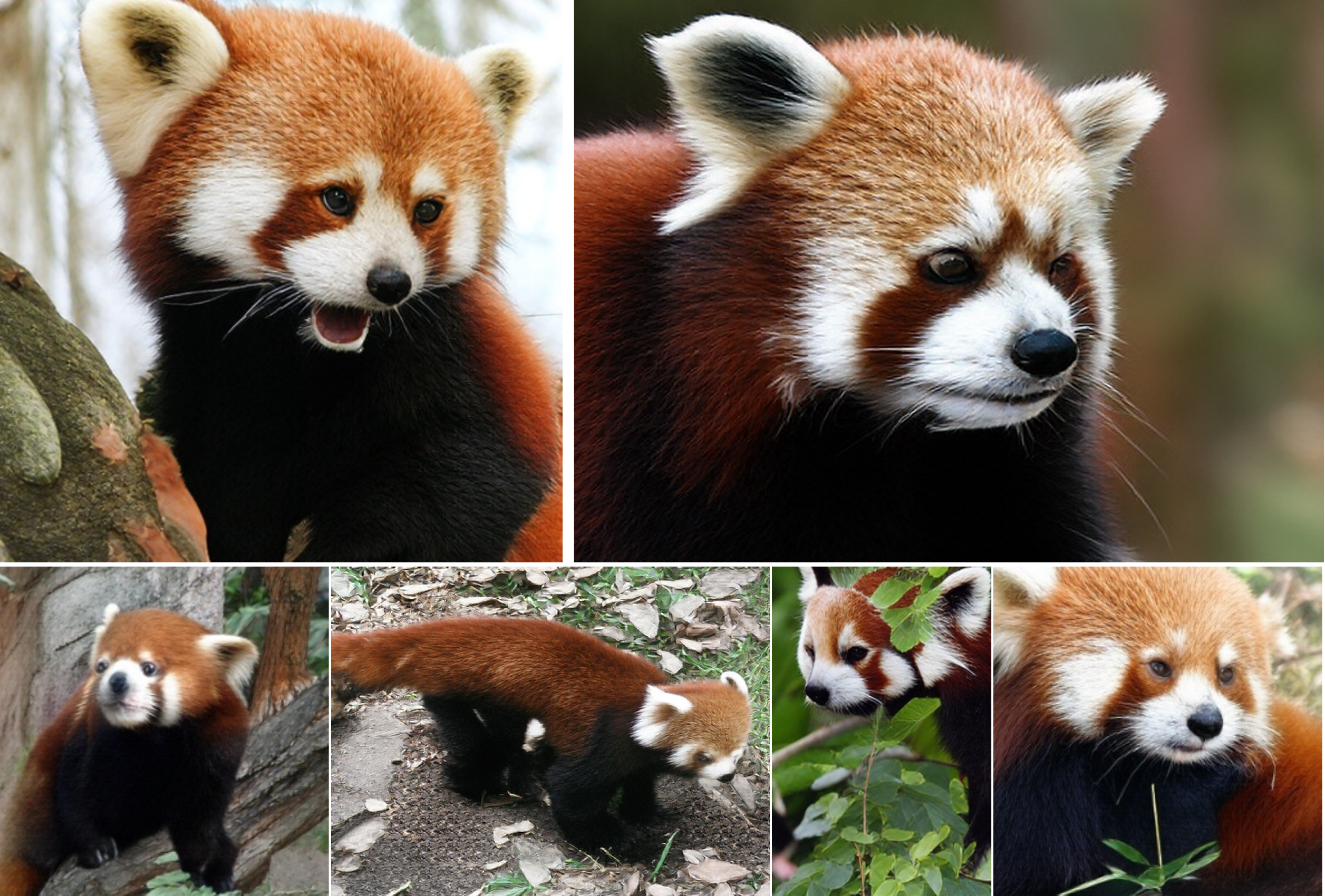}
    \caption{\textbf{Class 387.} Qualitative examples of generated images using \tokenvibegen\ (XXL/64) on arbitrary resolutions.
    }
    \label{fig:class_387}
\end{figure*}

\begin{figure*}[t]
    \centering
    \includegraphics[width=\linewidth]{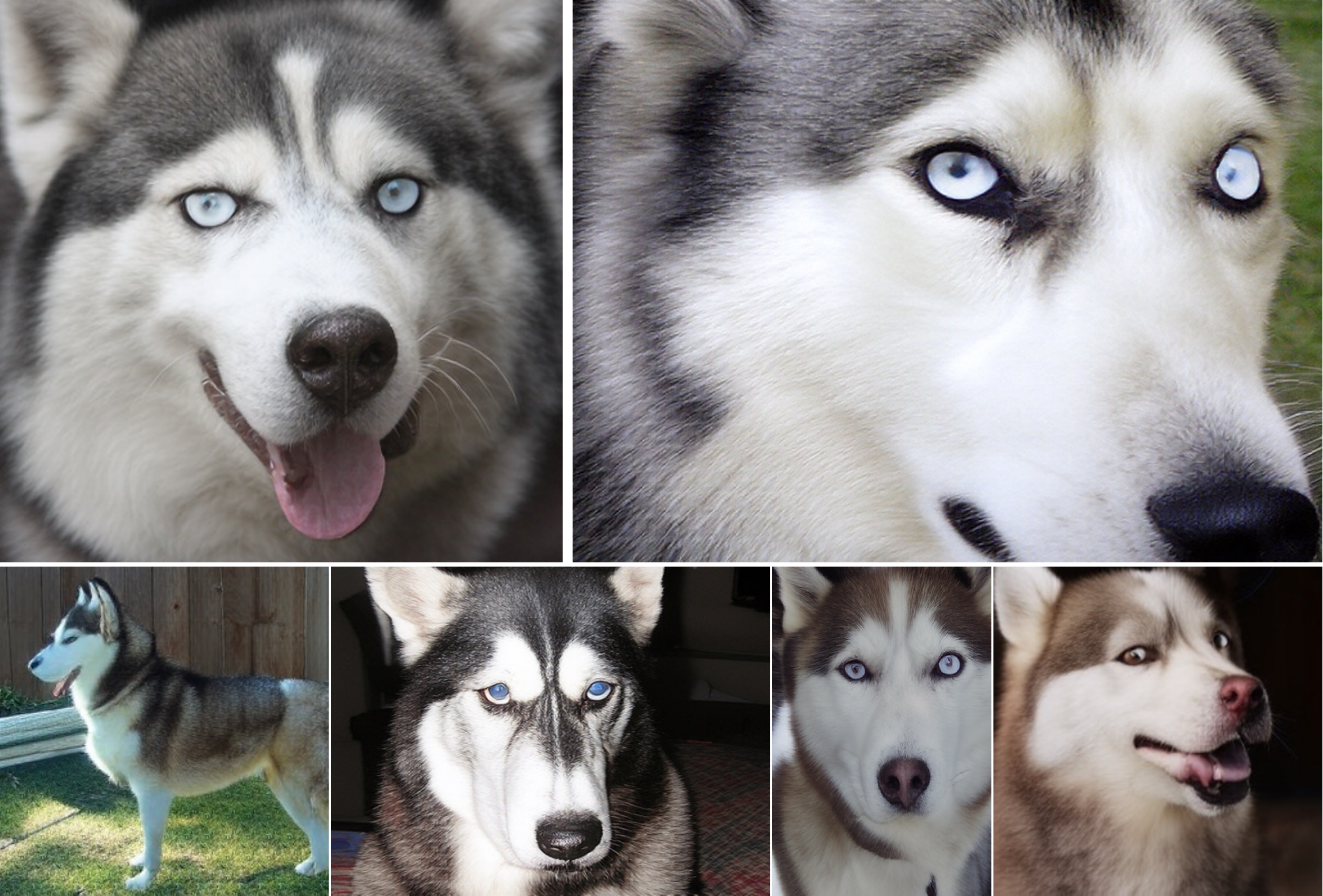}
    \caption{\textbf{Class 250.} Qualitative examples of generated images using \tokenvibegen\ (XXL/64) on arbitrary resolutions.
    }
    \label{fig:class_250}
\end{figure*}

\begin{figure*}[t]
    \centering
    \includegraphics[width=\linewidth]{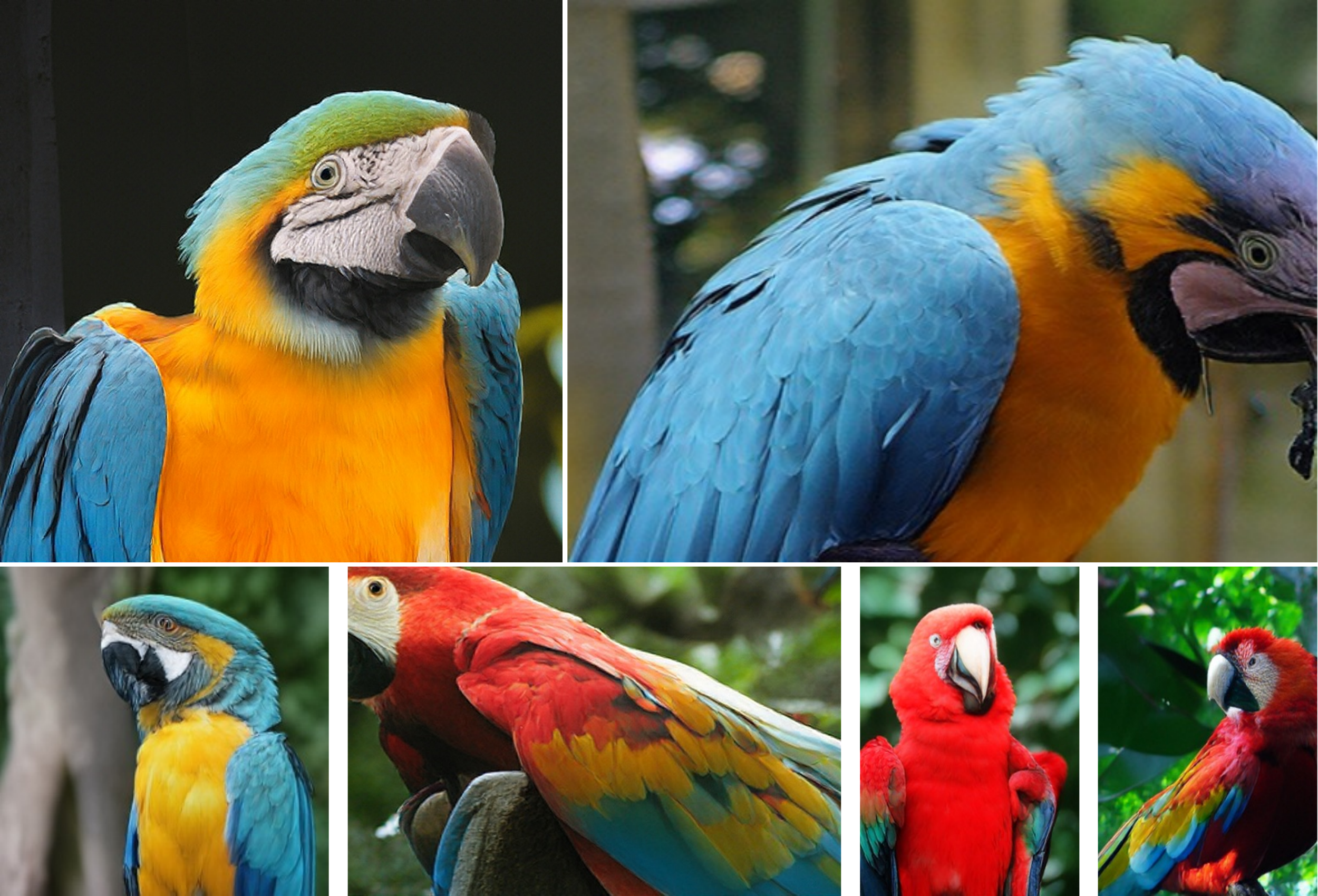}
    \caption{\textbf{Class 88.} Qualitative examples of generated images using \tokenvibegen\ (XXL/64) on arbitrary resolutions.
    }
    \label{fig:class_88}
\end{figure*}

\begin{figure*}[t]
    \centering
    \includegraphics[width=\linewidth]{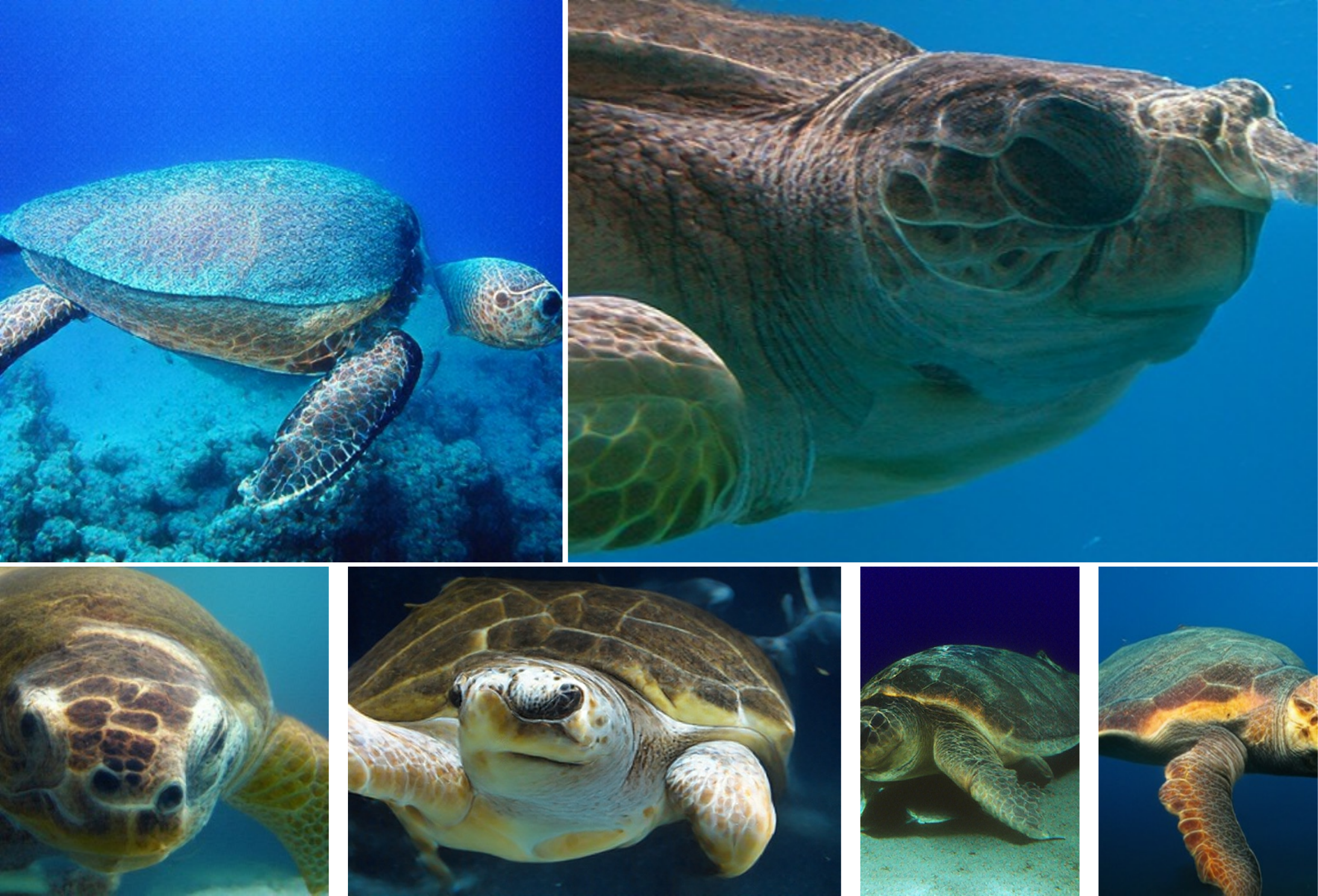}
    \caption{\textbf{Class 33.} Qualitative examples of generated images using \tokenvibegen\ (XXL/64) on arbitrary resolutions.
    }
    \label{fig:class_33}
\end{figure*}

Figures~\ref{fig:class_985}, \ref{fig:class_980}, \ref{fig:class_387}, \ref{fig:class_250}, \ref{fig:class_88}, and \ref{fig:class_33} illustrate samples from six ImageNet1k classes generated by \tokenvibegen\texttt{-(XXL/64)} at random resolutions. \tokenvibegen\ consistently produces high-resolution, variable-aspect-ratio images using only $64$ tokens. Some crops or truncations may appear; we attribute these artifacts to the randomized cropping used during AR training.

\section{Limitations and Future Works}
Our experiments focus on ImageNet-1k and class-conditional generation; extending to text-to-image, open-vocabulary settings, and video is an important next step. 
As a resolution \emph{generalist} model, \tokenvibegen\ can trail single-resolution specialists at exactly $256^2$/$512^2$ under the same budget. Further scaling, longer training, and stronger data augmentation may reduce this gap. 
Methodologically, exploring other AR variants (\eg randomized orderings, scale-wise training), alternative quantization schemes, and larger pretraining corpora could improve quality. Finally, applying resolution-agnostic tokenization to unified multimodal modeling (\eg image–text–video) is a promising direction for significantly efficient production-grade generative systems.

\end{document}